\documentclass[acmtog,authorversion,nonacm]{acmart}
\AtBeginDocument{%
  \providecommand\BibTeX{{%
    \normalfont B\kern-0.5em{\scshape i\kern-0.25em b}\kern-0.8em\TeX}}}

\setcopyright{none}
\copyrightyear{2024}
\acmYear{2024}
\acmDOI{XXXXXXX.XXXXXXX}

\usepackage{booktabs} 
\usepackage{amsmath}
\acmJournal{TOG}

\usepackage{algpseudocode}
\usepackage{xcolor}
\usepackage{multirow}
\usepackage{array}
\usepackage{graphicx}
\usepackage{booktabs} 
\usepackage{enumitem}
\usepackage{epsfig}
\usepackage{graphicx}
\usepackage{soul}
\usepackage{multirow}
\usepackage{booktabs}
\usepackage{wrapfig}
\usepackage{footnote}
\usepackage{subcaption}
\usepackage{booktabs,tabularx}
\usepackage{cleveref}
\usepackage{xcolor}
\usepackage[ruled,vlined]{algorithm2e}
\usepackage{tcolorbox}

\SetKwSty{simple}
\usepackage[linewidth=1pt]{mdframed}

\makeatletter
\newcommand*\smallE{\mathds{E}} 
\newcommand*\yo@bigE[2]{\text{#2\raisebox{-#1ex}{$\smallE$}}}
\DeclareRobustCommand*\bigE{\mathop{\mathchoice
  {\yo@bigE{0.5}{\larger\larger}} 
  {\yo@bigE{0.4}{\larger\larger}} 
  {\yo@bigE{0.3}{\larger\larger}} 
  {\text{\scalebox{1.0}{\raisebox{-0.5ex}{$\smallE$}}}} 
  }\displaylimits}
\makeatother

\newenvironment{packed_enumerate}
{\begin{enumerate}
    \vspace{-\topsep}
    \setlength{\itemsep}{1pt}
    \setlength{\parskip}{0pt}
    \setlength{\parsep}{0pt}
}{\end{enumerate}}

\newcommand{\methodpropfull}[1]{\textsc{Analytical Edit Propagation}\xspace}
\newcommand{\methodprop}[1]{\textsc{AEP}\xspace}

\newcommand{\progfull}[1]{\textsc{Analytical Shape Editing Programs}\xspace}
\newcommand{\prog}[1]{\textsc{ShapeAE}\xspace}

\newcommand{\rone}[1][1]{\textsc{\color{blue}R1}\xspace}
\newcommand{\rtwo}[1][1]{\textsc{\color{green}R2}\xspace}
\newcommand{\rthree}[1][1]{\textsc{\color{red}R3}\xspace}

\begin{document}

\title{ParSEL: Parameterized Shape Editing with Language}

\begin{abstract}

The ability to edit 3D assets from natural language presents a compelling paradigm to aid in the democratization of 3D content creation.
However, while natural language is often effective at communicating general intent, it is poorly suited for specifying precise manipulation.
To address this gap, we introduce \textsc{ParSEL}, a system that enables controllable editing of high-quality 3D assets from natural language. 
Given a segmented 3D mesh and an editing request, \textsc{ParSEL} produces a \textit{parameterized} editing program.
Adjusting the program parameters allows users to explore shape variations with a precise control over the magnitudes of edits.
To infer editing programs which align with an input edit request, we leverage the abilities of large-language models (LLMs).
However, while we find that LLMs excel at identifying initial edit operations, they often fail to infer complete editing programs, and produce outputs that violate shape semantics.
To overcome this issue, we introduce \textsc{Analytical Edit Propagation (AEP)}, an algorithm which extends a seed edit with additional operations until a complete editing program has been formed. 
Unlike prior methods, \textsc{AEP} searches for analytical editing operations compatible with a range of possible user edits through the integration of computer algebra systems for geometric analysis.
Experimentally we demonstrate \textsc{ParSEL}'s effectiveness in enabling controllable editing of 3D objects through natural language requests over alternative system designs.

\end{abstract}

\author{Aditya Ganeshan}
\email{adityaganeshan@gmail.com}
\affiliation{%
    \institution{Brown University}
    \country{USA}
}

\author{Ryan Y Huang}
\email{ryan_y_huang@brown.edu}
\affiliation{%
    \institution{Brown University}
    \country{USA}
}
\author{Xianghao Xu}
\email{xianghao_xu@brown.edu}
\affiliation{%
    \institution{Brown University}
    \country{USA}
}
\author{R. Kenny Jones}
\email{russell_jones@brown.edu}
\affiliation{%
    \institution{Brown University}
    \country{USA}
}
\author{Daniel Ritchie}
\email{daniel_ritchie@brown.edu}
\affiliation{%
    \institution{Brown University}
    \country{USA}
}
\renewcommand{\shortauthors}{Ganeshan and Huang, et al.}

\ccsdesc[500]{Computing methodologies~Computer graphics}
\ccsdesc[500]{Computing methodologies~Neural networks}
\ccsdesc[500]{Computing methodologies~Natural language generation}

\keywords{Shape Editing, Parametric Editing, Large Language Models, Computer Algebra Systems, Neuro-Symbolic Methods, Program Synthesis}

\setlength{\abovedisplayskip}{3pt}
\setlength{\belowdisplayskip}{3pt}
\begin{teaserfigure}
    \centering
    \includegraphics[width=1.0\linewidth]{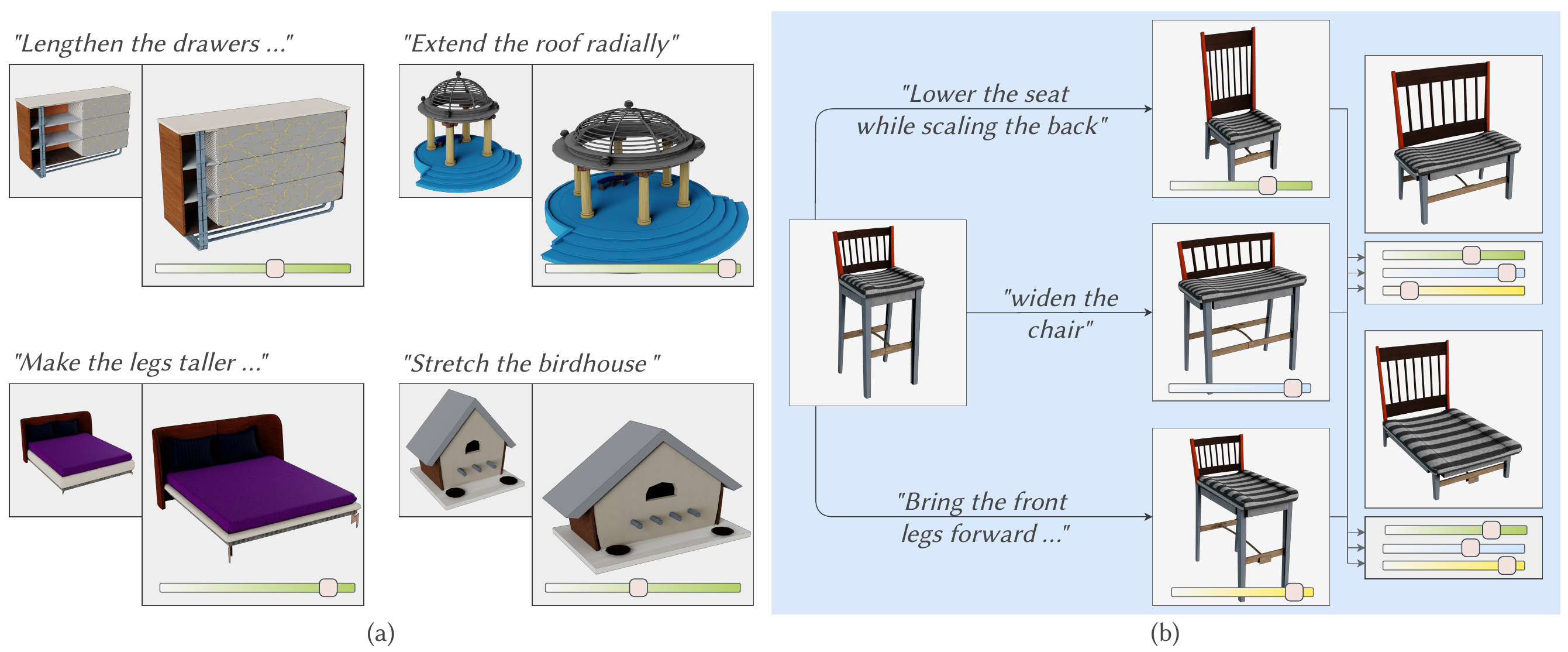}
    \vspace{-1.25em}
    \caption{
    We introduce \textsc{ParSEL}, a system that enables \textit{controllable} editing of 3D assets with natural language.
    (a) Each subplot shows an input 3D asset (left), edit request (top) and the parametric editing capability provided by  \textsc{ParSEL} (right).
    (b) The parametric edits produced by \textsc{ParSEL} are composable, allowing users to explore shape variations of \textit{non-parametric} models as seamlessly as they would with parametric models. 
    }
    \label{fig:teaser}
\end{teaserfigure}
\maketitle

\section{Introduction}

Creating high quality 3D assets is a labour intensive task, requiring years of training and experience.
The ability to edit existing high quality 3D assets to create new assets greatly lowers this barrier.
However, the process of manually modifying a 3D object, e.g. by adjusting individual vertices and/or faces, is often tedious.
To address this challenge, a number of efforts have explored how to design more use-friendly and intuitive shape editing tools ~\cite{geo_snap, yumer_semantic}.

Following rapid advancements in the field of Natural Language Processing, a wealth of recent methods have explored techniques for editing 3D assets from natural language~\cite{shapewalk, achlioptas2023shapetalk, Gao_2023_SIGGRAPH, text2mesh}. 
This proffered paradigm is alluring from the perspective of accessibility; specifying edit intent through natural language requires minimal tool-specific training. 
So far, such natural language based editing systems have shown promising results for retexurizing 3D assets~\cite{aladdin}, stylizing 3D shapes~\cite{text2mesh} and even modifying shape geometry~\cite{achlioptas2023shapetalk}. 

However, editing shapes in a controlled manner with language is challenging,
particularly when performing \textit{geometric edits}, i.e. edits involving sub-part manipulation and spatial rearrangement.
For instance, consider a scenario where a user intends to widen a chair. 
Natural language can be used to effectively communicate \textit{how} to edit: ``scale the seat, reposition the legs and add more back slats''. 
Yet, it is difficult to convey \textit{how much} to edit with natural language.
Terms such as “moderately widen,” “greatly widen” are inherently subjective, and numerical specification (``0.2 units'') often may not align with the object’s scale.
This leads us to a critical insight: \textit{while natural language can effectively convey the edit intent, it is poorly suited for conveying the edit magnitude}.
As a result, systems that depend exclusively on language for shape editing are often challenging to use in practice.

To facilitate \textit{controllable} editing of 3D assets using natural language, we introduce \textsc{ParSEL: Parameterized Shape Editing with Language}, a system where users define "how" to edit with language and "how much" with adjustable parameters. 
Drawing inspiration from parametric modeling systems, our approach merges the intuitiveness of language with the precision of parametric control. 
Under this paradigm, users can seamlessly explore a family of shape variations by adjusting parameters until they find an edit that matches their intended magnitude. 
In Figure~\ref{fig:teaser} (a), we showcase qualitative examples of editing various 3D shapes using our system.

\textsc{ParSEL} takes as input a semantically labeled 3D mesh and a user edit request.
As output, it exposes an adjustable parameter which the user can interact with to flexibly explore shape variations.
We geometrically realize these edits by propagating part-level bounding proxy deformations to the underlying mesh geometry with cage-based deformation methods~\cite{harm_coords}, while simultaneously adapting part-level symmetry groups.
Critically, to ensure that the geometry remains consistent under a range of parameter variations, we represent all edit operations as closed-form analytical functions of the adjustable control parameters. 
In order to create such parameterized editing functions, we introduce a custom domain specific language.
Programs in our DSL specify \textit{how} to edit the shape with numerical parameters, and \textit{how much} to edit with algebraic expressions of the control parameters. 
Our DSL offers multiple benefits, including the ability to provide a fluid (solver-less) edit exploration experience to the users, even on consumer laptops. 

\textsc{ParSEL} converts language-based edit requests into editing programs with a module that integrates large-language models (LLMs) with the algebraic reasoning capabilities of computer algebra systems (CAS).
We found that this coupled neurosymbolic approach is necessitated as current state-of-the-art LLMs struggle to directly infer editing programs from input edit requests.
Due to their poor geometric reasoning capabilities, LLMs often fail to infer the appropriate adjustments required for multiple input shape parts, resulting in editing programs which produce inconsistent outputs (we explore this phenomenon further in Section~\ref{sec:task}).

To overcome this limitation, we present an algorithm that extends partial editing programs with additional operators from our DSL by considering the geometric relations between object parts. We term this technique \textsc{Analytical Edit Propagation (AEP)}. While similar in spirit to prior edit-propagation methods~\cite{comp_prop}, previous techniques explicitly optimize part modifications in response to a user edit. In contrast, we \textit{search} for analytical editing functions compatible with a range of possible user edits using a sophisticated CAS solver~\cite{sympy}. 
Furthermore, by integrating LLMs with \textsc{AEP}, we address a key limitation of all edit propagation-based editing methods~\cite{iwires, comp_prop} - the requirement to manual adjust the shape’s structure to support different editing intents. By leveraging LLMs, we dynamically modify the shape’s structure based on the edit requests, thus alleviating the need for manual adjustments.

We evaluate \textsc{ParSEL}'s ability to edit 3D objects from language by sourcing \textit{(shape, edit request)} pairs from assets in CoMPaT3D++~\cite{slim2023_3dcompatplus} and PartNet~\cite{partnet}. 
On pairs from CoMPaT3D++, we compare \textsc{ParSEL}'s hybrid approach for producing an editing program against two alternative formulations: (i) asking an LLM to directly author the program, and 
(ii) using the LLM and \textsc{AEP} to infer full program without dynamically altering the shape structure.
We design a perceptual study to assess how well these variants accomplish the intended task and find that participants greatly prefer the edits produced by \textsc{ParSEL}.
To provide further analysis on the design decisions behind \textsc{ParSEL}, we create expert-designed editing programs for PartNet pairs. Treating these manual annotations as `ground-truth', we perform isolated ablation experiments on the LLM prompting workflow and the CAS solver. We also investigate two extensions of our method. First, we demonstrate how \textsc{ParSEL} can aid in creating (approximate) parametric models of non-parametric assets by composing a series of edit requests (Figure~\ref{fig:teaser}, (b)). Then, we explore how \textsc{ParSEL} can produce a multitude of shape variations from a single user editing request.

In summary, we make the following contributions: 
\begin{packed_enumerate}
    \item We introduce \textsc{ParSEL: Parameterized Shape Editing with Language}, a novel shape editing system which combines the intuitiveness of natural language with the precision of parametric control.
    \item We design a neurosymbolic module which couples a LLM prompting workflow with CAS solvers to translate edits expressed in natural language into shape editing programs.
    \item To successfully solve the above inference task, we introduce \textsc{Analytical Edit Propagation} an algorithm to search for \textit{analytical} editing functions to extend partial editing programs by considering inter-part geometric relationships.
\end{packed_enumerate}

Code of our system  will be open-sourced upon publication.

\section{Related Work}
In this section, we review prior works in 3D shape editing, focusing on analyze-and-edit techniques and language-based editing methods. We also compare our approach to existing parameterized editing methods, highlighting key differences. Finally, we discuss recent approaches which leverage large language models (LLMs) for visual program synthesis.

\textit{Analyse-and-edit approaches:}
Shape editing using space deformation has a long history in computer graphics, traditionally employing simpler control objects like cages to manipulate underlying meshes~\cite{Sederberg, Coquillart}. 
This method evolved significantly with the development of \textit{analyze-and-edit} techniques~\cite{orig_wires, Sumner, iwires, comp_prop}, which provide better control by aligning control objects more closely with the shape's structure. 
Consequently, part-decomposition and symmetry group-aware control objects are constructed to facilitate structure preservation during editing~\cite{pattern_docker, wang2011}.

Various approaches have been proposed within this paradigm, tailored for tasks such as resizing non-homogeneous shapes~\cite{resize_old}, editing articulated objects~\cite{joint_deform}, modifying architectural scenes~\cite{architecture, architecture_2}, and manipulating 2D SVG patterns~\cite{GuerreroEtAl:PATEX:2016}. 
Please refer to~\cite{niloy_survey} for a more complete review.
Typically, these methods require the user to perform an initial edit, usually though a visual user interface, like dragging a point, after which the system propagates corrective adjustments throughout the shape via computationally intensive numerical optimization. 
This optimization process must be repeated even for simple adjustments in the magnitude of the intended edit. 
In contrast, our technique precomputes analytical editing programs that align with the user's intent, enabling real-time adjustments to the edit's magnitude. 
Additionally, previous methods often support only a subset of the edits we offer, necessitate manual adjustments to the shape's structure (such as deleting symmetry relations), and do not facilitate editing through natural language.

\textit{Language based shape editing:}
The advancement in text-to-image modeling capabilities has led to many language-based editing systems~\cite{lin2023text, brooks2022instructpix2pix}. 
Works such as~\cite{wang2023mdp, Kim_2022_CVPR} have introduced tools to edit images with natural language, although these edits are often limited to stylization. 
Due to the weak spatial and geometric understanding of text-to-image models, such approaches typically fail at geometric editing.
This trend extends to language-based 3D shape editing, where tools for editing asset textures~\cite{aladdin}, stylizing a mesh, including its geometry~\cite{text2mesh, Gao_2023_SIGGRAPH} have seen success. 
However, approaches using text-to-image models are yet to demonstrate impressive performance for \textit{geometric} editing of 3D shapes.

Another line of research~\cite{shapewalk, achlioptas2023shapetalk} has focused on training shape-editing models on datasets comprising 3D shapes paired with editing requests and the corresponding edited shapes. 
While these results are promising, they have not scaled to the quality required for real-world 3D asset editing, often manipulating only low-resolution point clouds~\cite{achlioptas2023shapetalk} or implicit functions~\cite{shapewalk}. 
Additionally, these approaches struggle with the inherent challenge of using natural language to specify an edit's magnitude. 
Furthermore, while end-to-end modeling allows these models to handle a wide variety of edits, this flexibility often results in entangled edits, where parts that should remain unchanged are inadvertently modified. 
In contrast, our work avoids these drawbacks by inferring analytical editing programs aligned with the user's intent, enabling precise and disentangled geometric edits in real-time,
while eliding the requirement of a training dataset. 
We shown an example contrasting our approach to ShapeTalk~\cite{achlioptas2023shapetalk} in Figure~\ref{fig:shapetalk}, highlighting the discussed drawbacks.

\textit{Parameterized editing of Shapes:}
Lilicon~\cite{lillicon} introduced an SVG icon editing system that allows users to perform parameterized edits on SVG icons independently of the drawing's construction. 
Our editing system aims to offer similar functionality for 3D shapes by inferring parameterized editing programs based on the user's request. 
However, while Lilicon uses differential manipulation to maintain inter-part relationships, we instead search and utilize analytical edits that maintain these relationships.
Recently, coupling of parametric control with natural language has also been explored for image editing~\cite{guerrero2024texsliders, cheng2024learning}, though the edits are closer to stylization than geometric editing.

The need for high-level parametric control for shape editing has also been approached from a shape-abstraction perspective~\cite{jones2021shapeMOD, jones2023ShapeCoder}, aiming to discover program abstractions that allow users to edit shapes meaningfully with a few parameters. 
These data-driven methods, although effective at modeling shape variation within a dataset, may not capture a user's desired edit. 
Furthermore, such systems often provide unlabeled parameters, requiring users to experiment to understand each control's effect. 
In contrast, our system offers user controls that are \textit{directly} inferred based on the user's specified edit intent, providing more intuitive editing capabilities.

We also find an interesting contrast with approaches such as~\cite{reparam_cad, diff_cad, dag_amendments}, which propose techniques to make editing of parametric models easier, whereas our approach makes parameterized editing of \textit{non-parametric} models easier. 
One of our applications relates to inverse parametric modeling~\cite{aliaga2007style}. However, unlike prior approaches~\cite{bokeloh, aliaga2007style}, which create inverse parametric models strictly from geometric analysis, our approach leverages natural language to explore different procedural models for the same geometry.

\textit{Using LLMs for Visual Program Synthesis:}
The use of LLMs for visual tasks has proliferated in the recent years. Starting from earlier works employing LLMs for visualQA~\cite{surismenon2023vipergpt, visprog}, they have now been employed for image-editing~\cite{feng2023layoutgpt}, scene generation~\cite{yang2023holodeck, hu2024scenecraft}, inverse computer graphics~\cite{kulits2024igllm}, generating images~\cite{yamada2024l3go} and material modeling~\cite{huang2024blenderalchemy}. Though initial works such as~\cite{visprog} tried to rely purely on LLMs for their tasks, the subsequently winning model has been combining LLMs with domain specific structural-aware processing modules~\cite{huang2024blenderalchemy}. 
Our work also applies this formula but for shape editing.

\section{Overview}
\label{sec:overview}

\begin{figure*}
    \centering
    \includegraphics[width=1.0\linewidth]{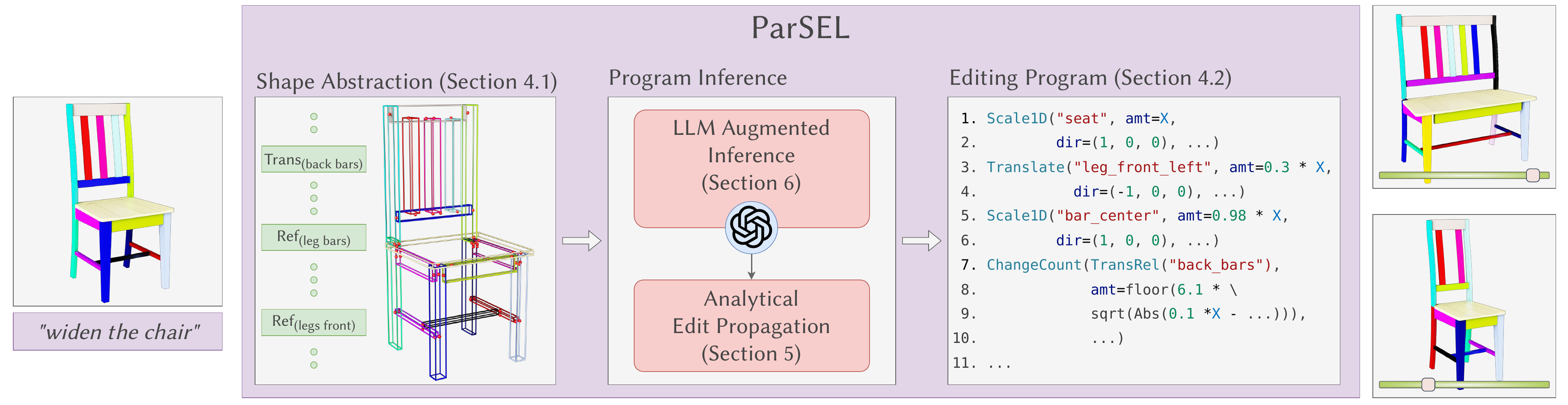}
    \caption{
\textbf{Overview}: Given a segmented 3D chair and an edit request to "widen the chair," we first convert the shape into a structured representation: hexahedrons and inter-part relations. This abstraction is illustrated on the left, with symmetry relations annotated on the left and attachment relations depicted with red points on the shape. Our neuro-symbolic approach (center) uses an LLM, to interpret the natural language input, and \textsc{Analytical Edit Propagation}, to perform geometric reasoning, in order to infer a \textit{parameterized} editing program (right) that aligns with the edit request. 
    }
    \vspace{-1.25em}
    \label{fig:method}
\end{figure*}

\textsc{ParSEL} takes 3D assets and natural language edit requests as input, exposing adjustable parameters to control the magnitude of the edit. 
Specifically, it processes an instance-level segmented 3D mesh paired with a natural language editing request to infer an \textit{editing program} parameterized by user-controlled parameters. Users can then interactively adjust the parameters to explore a wide array of shape variations reflecting different editing magnitudes.

Figure~\ref{fig:method} provides a schematic overview of our system. In Section~\ref{sec:editing_system}, we detail our approach for parameterized shape editing, outlining the shape representation and our custom Domain Specific Language (DSL) that facilitates parameterized shape editing. 
Our goal is to translate natural language edit requests into editing programs within our DSL that accurately reflect the user's intent. 
While Large Language Models (LLMs) offer a promising solution for this translation, they struggle with the complexity of composing editing programs, as these often involve multiple interdependent editing operations. 
Consequently, although LLMs can correctly identify the initial operation, known as the \textit{seed-edit}, they often fail to infer complete editing programs.

To address this limitation, in Section~\ref{sec:aep}, we introduce \textsc{Analytical Edit Propagation (AEP)}, which takes the \textit{seed-edit} inferred by an LLM and introduces additional editing operations to complete the editing program. This approach simplifies the task for the LLM, requiring it to infer only the initial seed edit. Additionally, by employing computer algebra systems for geometric reasoning, \textsc{AEP} ensures that the resulting editing programs have high geometric fidelity and respect essential shape features such as connectivity and symmetry.
Integrating LLMs with \textsc{AEP} significantly enhances our system's capability to meet user expectations by effectively bridging the gap between interpreting the primary edit and performing comprehensive shape adjustments.

Finally, Section~\ref{sec:llm} discusses our LLM Augmented Inference approach. 
A key limitation of all edit propagation methods~\cite{comp_prop, iwires} is the need for manual adjustments to the shape's structure to support different editing intents. 
This section introduces our solution to this challenge: leveraging LLMs to dynamically modify the shape's structure based on the edit requests. 
We end this section by briefly discussing our prompting workflow and the techniques we employ to boost the robustness and accuracy of LLM responses, ensuring reliable edits across diverse inputs.

\section{Parameterized Shape Editing}
\label{sec:editing_system}

In this section, we present our framework for enabling \textit{parameterized} editing of 3D shapes. 
First, we detail the shape representation employed in our system in Section~\ref{sec:shape}. Next, we describe our Domain Specific Language (DSL) for creating programs that facilitate parameterized shape editing in Section~\ref{sec:DSL}. 
Finally, in Section~\ref{sec:task}, we discuss the limitations of directly inferring complete editing programs in this DSL using LLMs, highlighting the need for \textsc{Analytical Edit Propagation (AEP)}.

\subsection{Structured Shape Abstraction}
\label{sec:shape}
Our system processes 3D meshes that are annotated at the instance level. We first transform these meshes into a structured representation to facilitate parametric editing. 
This representation is inspired by shape representations such as \textit{3D part graphs}~\cite{mo2019structedit} and \textit{Sym-Hierarchy}~\cite{wang2011}.

Specifically, we model the input shape $S$ as an undirected graph $G(N,E)$, where each node $n_i \in N$ corresponds to a semantically labeled mesh sub-part, and edges $e_i \in E$ denote inter-part geometric relations such as connectivity or symmetry. To preserve the detailed geometric features of the input mesh while simplifying the complexity of edits, each part $P_i$ is abstracted into a hexahedron $H_i$. This hexahedron, initialized with the part’s oriented bounding box, acts as the control cage for the underlying part-mesh. Edits made to the hexahedrons are then propagated to deform the vertices of the part mesh using harmonic coordinates~\cite{harm_coords}.

Our system support two types of inter-part relations, namely \textit{Symmetry Relation}, which models inter-part symmetry groups commonly found in man-made objects and \textit{Attachment Relation} which constrain the relative movement between parts, ensuring that edits do not violate the physical connections of the object.
Additionally, we automatically extend each part's label with verbal directional phrases such as "front" and "back" to capture its relative positioning among other instances with the same label. 
Figure~\ref{fig:method} illustrates the conversion of a chair model into this structured shape abstraction. 
We provide a more detailed overview in the supplementary materials.

\begin{figure*}
    \centering
    \includegraphics[width=1.0\linewidth]{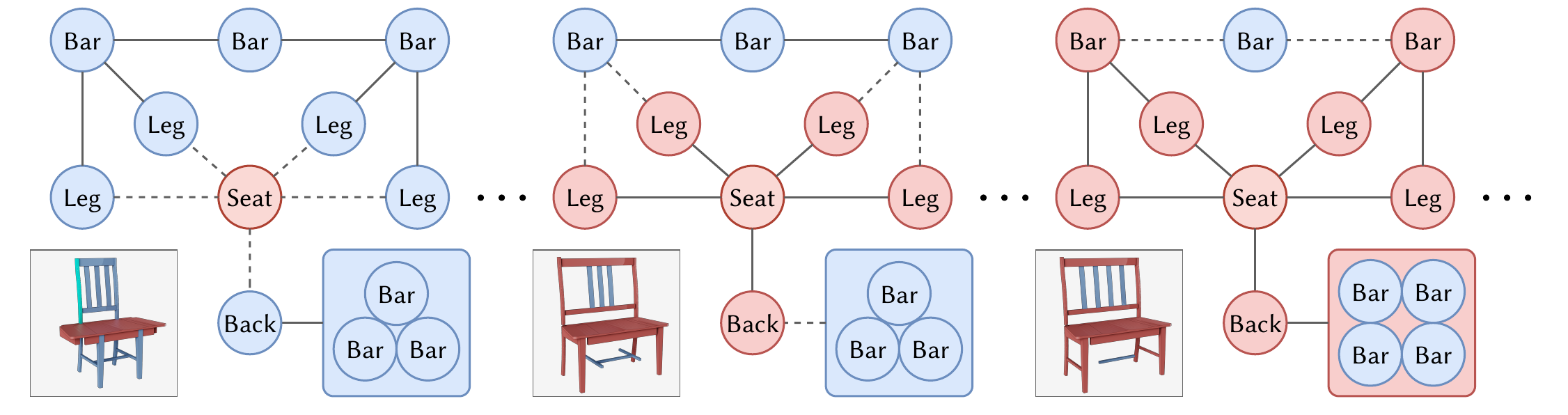}
    \caption{\textbf{Edit Propagation}: Starting with a \textit{seed-edit}, scaling the seat, new edits are incrementally introduced to rectify the broken relations. (left) Initially the seat-leg and seat-back attachments are broken. (middle) New edits, shifting the legs and scaling the back, are introduced to restore these broken relations. Consequently, the leg-bar and back-bar attachments are broken. This process continues until no relation remains broken, or all parts are edited. 
    }
    \vspace{-1.25em}
    \label{fig:prop}
\end{figure*}
\subsection{A Language for parameterized shape editing}
\label{sec:DSL}

In the preceding section, we introduced our structured shape representation, which is comprised of individual parts (abstracted as hexahedrons) and inter-part geometric relations.
Building upon this, we now present a Domain Specific Language (DSL) crafted to facilitate parametric editing of these components using user-controlled parameters.

Our DSL, enables common transformations on individual parts, including translation, rotation, scaling, and shearing. These can also be targeted at specific part features such as a face, edge, or corner. Additionally, the DSL supports modifications to symmetry group parameters, like the number of elements and their spacing.
A key feature of our DSL is the parameterization of editing operations via user-controlled parameters. This functionality is encapsulated with multiple 
atomic editing operators, collectively referred to as \textsc{EditOp}, and defined as follows:
\begin{equation*}
\textsc{EditOp}(\textsc{Operand}, \textsc{Amount}, \textsc{**params}),
\end{equation*}
where \textsc{EditOp} signifies the operation, \textsc{Operand} the target (part, feature, or symmetry relation), and \textsc{params} additional numerical parameters specific to each operation (e.g., a $\mathbb{R}^3$ vector for scaling operation's origin).
Central to these operators is $\textsc{Amount}$, the parameter dictating the magnitude of the edit. Unlike the other static parameters, $\textsc{Amount}$ is specified as a symbolic mathematical expressions of the user-controlled parameters, and dynamically evaluated as the user alters these parameters. As a result, adjusting the user-controlled parameter alters the edit magnitude, and subsequently helps explore shape variations.
We denote a sequence of these operations as a \textit{parameterized editing program}.
In Figure~\ref{fig:method} (c), we display a program designed to increase the width of a chair while proportionally adjusting other parts of the chair.

We design our DSL to achieve our goal of enabling \textit{controllable} shape editing.
We highlight a few pivotal aspects of our DSL:
(i) Unlike prior editing systems, our DSL handles a comprehensive range of operations. For example, ~\cite{bokeloh} only supports coupling of translation symmetries with part deformations, ~\cite{comp_prop} does not edit symmetry hierarchies, and ~\cite{wang2011} only supports editing of symmetry hierarchies.
(ii) The $\textsc{Amount}$ expression supports arbitrarily complex mathematical expressions, as demonstrated in the translation symmetry group edit in Figure~\ref{fig:method} (c). 
(iii) The ability to apply the transforms on part features facilitates non-affine transformations, such as tapering, which is useful in many edits (see  Figure~\ref{fig:qualitative} (c)), 
(iv) Each part's edits are independent of the state of other parts.  Consequently, these edits can be executed in parallel across all parts, enabling real-time parametric shape editing even on consumer-grade laptops. 
(v) The edits are composable; parts modified by one operation can seamlessly serve as inputs to subsequent operations.  This capability is explored further in Section~\ref{sec:proxydural}, where we demonstrate an exciting application of our system.
For a comprehensive discussion on the design and execution model of our DSL, please refer to our supplementary materials.

\subsection{Limitations of Direct LLM Inference}
\label{sec:task}

As we briefly discussed in Section~\ref{sec:overview}, inferring complete editing programs directly with LLMs often fails to produce results that align with the intended edits. 
LLMs typically succeed at interpreting primary edits explicitly requested—for example, when asked to `widen the chair's seat,' they correctly infer the edit operators to scale the seat. 
However, they often neglect necessary secondary edits crucial for maintaining geometric coherence, such as modifying related components like the chair's legs and back to accommodate the scaled seat. 
Additionally, even when LLMs identify the need for these secondary adjustments, they struggle to accurately infer the appropriate edit types and their magnitudes. 
As illustrated in Figure~\ref{fig:method} (right), editing programs can contain many operations that require careful adjustment of parameters to edit the shape cohesively. Consequently, the \textsc{Amount} parameter of different edit operations often involves complex mathematical expressions, making accurate inference challenging. As a result, we find that typically only the initial \textit{seed-edit} operation inferred by the LLM is reliable.

This limitation is overcome by \textsc{Analytical Edit Propagation (AEP)}, which performs the geometric analysis that LLMs cannot. Since the geometric relations driving these secondary edits are compatible with algebraic reasoning, \textsc{AEP} employs Computer Algebra System (CAS) solvers to identify the necessary secondary edits, thereby extending the editing program to better align with the user’s overall intent.

\section{Analytical Edit Propagation}
\label{sec:aep}

We now present our approach to extending a \textit{seed-edit} inferred by the LLM using CAS solvers for geometric reasoning. 
In Section~\ref{sec:prop}, we explain how edit propagation works and why parameterized editing operations require a different strategy.
Section~\ref{sec:shape_algebra} details our representation of edits, parts, and relations using algebraic expressions, resulting in parameterized constraints that the shape must satisfy. 
Finally, Section~\ref{sec:solve} describes how analytical solvers are employed to solve these parameterized constraints, thereby discovering appropriate new parameterized edits to extend the seed editing program.

\subsection{Edit Propagation}
\label{sec:prop}
When designing shape variations, maintaining the shape's functional and structural attributes is paramount.
We delineate these attributes through inter-part relations, as introduced in Section~\ref{sec:shape}. Thus, an adept editing program strives to \textit{preserve} these relations. 
Isolated part modification, a tendency of the naive strategy discussed in Section~\ref{sec:task}, generally result in the disruption of these inter-part relations. 
We now introduce a edit propagation algorithm which, starting with one or more seed edits, introduces additional edits to rectify the compromised relations.

As detailed in Section~\ref{sec:shape}, we model the input shape as an undirected graph $G(N, E)$, where each node $n \in N$ represents a subpart, and each edge $e \in E$ denotes an inter-part relation. 
Editing a node, such as widening a chair’s seat, can break the edges connected to this node - for example, it might break the attachment relation between the seat and the legs. 
Edit propagation is then initiated to restore the broken edges by applying corrective edits to other, previously unaltered, nodes within the graph.  
Importantly, these corrective edits are derived by considering broken relations with \textit{edited} parts only.
Consequently, newly introduced edits may subsequently break additional edges, necessitating further edits. 
This iterative process continues, introducing necessary edits until no edges remain broken or until all nodes have been edited. 
In Figure~\ref{fig:prop}, we demonstrate edit propagation in action, incrementally introducing new edits which eventually result in a cohesive edit of the input shape.

As our seed edits are \textit{parameterized} with user-controlled parameters, they model a range of potential instantiated seed edits. 
While traditional edit propagation techniques~\cite{comp_prop} can be applied to the dynamically evaluated instances of these seed edits, such application results in a laggy user experience. 
This lag arises because these techniques involve multiple computationally intensive numerical optimization iterations, dependent on the graph's size.
Instead, we propose to search for new parameterized edits that restore and preserve inter-part relations across the range of control parameters. Although this approach requires a computationally intensive search initially, it ensures a smooth, real-time user experience when the control parameters are later adjusted. 

To search for parameterized edits, we employ a sophisticated Computer Algebra Systems (CAS)~\cite{sympy}.
By representing parts, relations, and edits as algebraic expressions, we enable the use of CAS solvers to efficiently discover analytical solutions to the algebraic constraints dictated by the inter-part relations. As we will see later, this plays a critical role in finding edit parameterizations that preserve relations across the control parameter's range.

\subsection{Expressing the Shape Algebraically}
\label{sec:shape_algebra}
For simplicity, we assume that there is a single control parameter $x$ and refer to its range as the input range.  
First, we present the algebraic form of the part hexahedrons $H_i$. Then, we present how all inter-part relations are expressed as algebraic constraints on the the hexahedrons.

Each part hexahedron consists of $8$ 3D points, and is represented as a matrix of size $(8, 3)$.
A hexahedron under no edit is expressed as a constant function $H_i(x) = H_i^0 \in \mathbb{R}^{\{8, 3\}}$, where $H_i^0$ contains real-valued entries.
When a parameterized edit is applied on $H_i$, its functional form can be derived based on the type of edit.
For instance, under an edit $\textsc{Translate}(H_i, x, \hat{v})$, $H_i$ can be expressed as $H_i(x) = H_i^0 + x \cdot \hat{v}$, where $\hat{v}$ represents the direction of translation. 
Similar algebraic expressions can be prescribed for all the atomic editing operators in our DSL,  allowing us to express all the hexahedrons as algebraic functions of the control parameter $x$.
For reference, we tabulate the algebraic form of all the operators in the supplementary.

For modeling the relations algebraically, we separately handle the symmetry and attachment relations. 
Given hexahedrons $H_i$ and $H_j$ under a symmetry group $G(T)$, where $T$ is the transformation under which the symmetry is held, each symmetry relation can be expressed as a constraint $||H_i - T(H_j)||_\infty <\delta$. Similarly, attachment relations are expressed as $||a - b||_\infty <\delta$, where $a$ and $b$ are points in $H_i$ and $H_j$ respectively which form the attachment.  
By modeling $a$ and $b$ with harmonic coordinates, we can rewrite this constraint in terms of the hexahedron: $||M_a H_i - M_b H_j||_\infty <\delta$ where $M_a$ and $M_b$ are the harmonic coordinates of points $a$ and $b$ respectively. 
Since the hexahedrons $H_i$ are parameterized by $x$, these constraints are also parameterized by $x$ as can be expressed as $C_i(x)$.
Now, we can identify if a relation will be broken under the parameterized edits by checking if the corresponding constraints are maintained across the input range. More formally, we denote that a constraint is held by $SAT(C)$, and 
\begin{equation}
SAT(C) \implies ||C_i(x)||_\infty <\delta \forall x \in [0, \tau],
\end{equation}
where $[0, \tau]$ is the input range.
Ideally, $SAT(C)$ should be checked analytically, considering the functional form of $C(x)$. However, this can be computationally expensive. Therefore, we instead numerically check if a constraint is held, by evaluating it over random values of $x$ sampled uniformly across the input range.

\subsection{Analytical Edit Solver} 
\label{sec:solve}
Now, given a set of broken relations, we are tasked with introducing new parameterized edits which restores and preserves these relations across the input range. 
For a single part $P_i$, this problem can be written as finding an edit $E^*$ for $P_i$ such that all the constraints on the part are held: 
\begin{gather}
    \text{Find } E^* \text{ s.t. } SAT(E, \mathbb{C}) \\ 
    SAT(E, \mathbb{C}) \implies \{||C_i(x)||_\infty < \delta |\forall C_i \in \mathbb{C}, \forall x \in (0, \tau) \},
\end{gather}
where $\mathbb{C}$ is the set of all constraints derived from the broken relations of the part.

Restoring symmetry group relations is quite simple, as it has a well defined analytical solution. Given $H_i$ and $H_j$ under a symmetry group $G(T)$, where $H_i$ has an edit $E$, we can introduce a new edit $E^* = T(E)$, on $H_j$ such that the symmetry group is preserved. For example, to restore reflection symmetry between two parts with one of them under a parameterized translation, we can introduce a similarly parameterized translation on the other part with its translation direction reflected about the reflection plane. Following~\cite{comp_prop}, we prioritize symmetry relation over the attachment relations, and restore them before searching for attachment preserving edits. 

\begin{figure}
    \centering
    \includegraphics[width=1.0\linewidth]{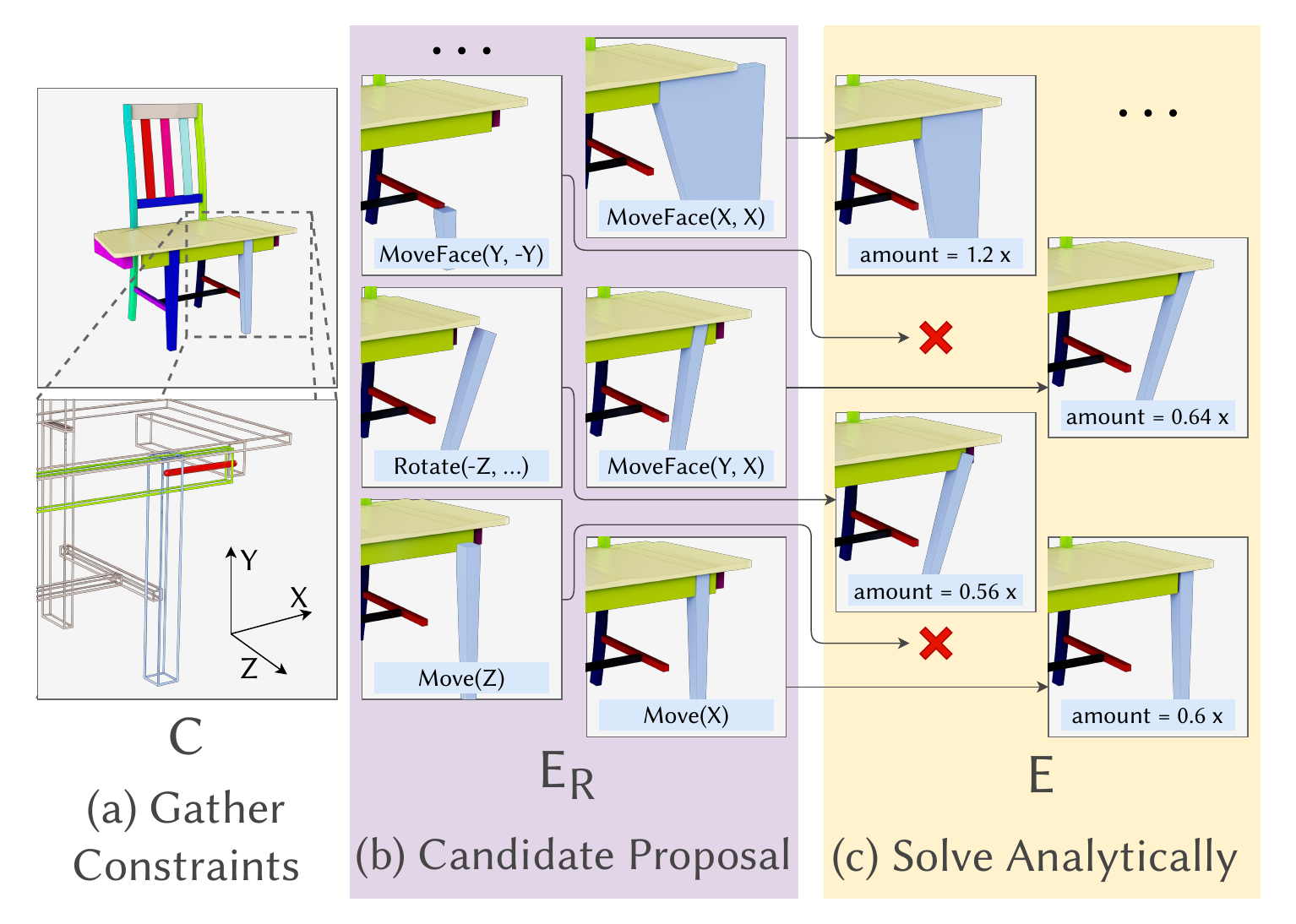}
    \caption{
    \textbf{Searching for edits}: Which edit operations could we use for the chair leg part? 
    (a) First, broken constraints $\mathbb{C}$ (highlighted with a red line) are detected, denoting the relations to be fixed.
    (b) We then sample parameterizations of the edit operators in our DSL to create candidate edits $\mathbb{E}_R$. 
    (c) Using CAS solvers, we search for $\textsc{Amount}$ expression for each edit candidate that satisfies the constraints $\mathbb{C}$, resulting in the set of valid edits $\mathbb{E}$.
    }
    \vspace{-1.25em}
    \label{fig:search}
\end{figure}

In contrast, restoring attachment relations is uniquely challenging. The set of constraints derived from attachment relations can be of arbitrary size (depending on the number of attachments relation of the part), without well defined analytical solutions. Therefore, we need to \textit{search} for edits that can satisfy all the constraints. This entails searching for (i) the correct type of edit, with (ii) the correct edit-specific parameters \textsc{params} which are defined over $\mathbb{R}^n$, and (iii) the correct parameterization \textsc{Amount} which can be an arbitrary algebraic expression of the the control parameter $x$. 
Naively performing this search is infeasible. We therefore employ two techniques to address this challenge. As we detail ahead, our solution relies on smartly sampling assignments of \textsc{params}, and using CAS solvers to infer feasible \textsc{Amount} expressions.
To aid exposition, we continue with the example of widening a chair seat, with an accompanying illustration in Figure~\ref{fig:search}. 
As shown in Figure~\ref{fig:search} (a), as the seat is widened, the attachment relation between the legs and the seat is broken. We are now tasked with finding a suitable edit for the leg to restore this attachment relation.

Each edit has numeric parameters \textsc{params} which can take arbitrary values in $\mathbb{R}^n$.
Instead of searching across all possible \textsc{params}, we search over a smaller subset of feasible \textsc{params} assignments. 
The feasible \textsc{params} assigments are created using the hexahedron features such as  its face-center, vertices, and local axis directions. 
We enumerate over the (applicable) DSL edit operators, and exhaustively sample feasible assignments of $\textsc{param}$ to create $\mathbb{E}_R$ a set of feasible edit candidates. In Figure~\ref{fig:search} (b), we depict the candidate edits from $\mathbb{E}_R$. 
Our goal is to now enumerate through the edit candidates, and ascertain if it can satisfy the constraints. 

To check if a edit candidate can satisfy the constraints, we must specify its \textsc{Amount}. 
Note that \textsc{Amount} is a algebraic expression and needs to be set s.t. it satisfies the constraints for all values of $x$ in the input range.
Therefore, we set \textsc{Amount} to be an arbitrary function $f(x)$ and state this task more formally as: 

\begin{equation}
    \text{Find } f(x) \text{ s.t. } SAT\big(E_{\textsc{Amount}=f(x)}, \mathbb{C} \big), E \in \mathbb{E}_R
\label{eq:analytic}
\end{equation}
Here, we leverage the CAS solvers to search for \textit{analytical} solutions for $f(x)$. 
Note that since we can have multiple constraints on a given part, the constraint set $\mathbb{C}$ can be a mixed set of equations - which CAS solvers may fail to find solutions for. Therefore, we instead create a set of feasible analytical solutions $\mathbb{F}$ by solving each constraint independently (i.e. by solving $f(x)$ s.t.  $||C_i(x)|| = 0$). \textit{Analytical} solutions in $\mathbb{F}$ which satisfy all constraints (by numerical check) are then accepted as solutions to equation~\ref{eq:analytic}.
We additionally note that solving equation~\ref{eq:analytic} for a given edit $E$ is ``embarrasingly'' parallel, allowing us to search solutions for different edits $E \in \mathbb{E}_R$ in parallel.

With the \textit{analytical} solutions, we form a set of $\mathbb{E}$ containing edits which restore all the constraints $\mathbb{E} = \{SAT(E, \mathbb{C}) | E \in E_R\}$. This set is depicted in Figure~\ref{fig:search} (c). 
Though all the edit candidates in $\mathbb{E}$ restore the \textit{currently} broken relations, each may cause a different set of relations to be broken.
Therefore, we must carefully select the edit candidate.
Taking inspiration from prior work~\cite{iwires}, we design a simple selection criterion for selecting the most suitable edit from $\mathbb{E}$, which considers the ARAP deformation energy~\cite{ARAP_modeling} (lower is better)
and the number of intrinsic symmetry planes of the edit (higher is better). 
We provide additional details in the supplementary.

\subsubsection{Improving Solver Robustness}
\label{sec:aep_ext}
Though the technique presented above succeeds in a majority of analytical edit propagation steps, we found that it can sometimes fails to find good edits. We now detail two features which improve the quality of solutions found by the Solver.
As we experimentally verify later in Section~\ref{sec:ablation}, these features noticeably improve the program quality. 

\noindent
\textbf{Extending the Candidate Set} 
The set $\mathbb{E_R}$ of edit candidates can sometimes fail to contain \textsc{param} assignments which can successfully satisfy all the constraints. When none of the edits in $\mathbb{E}_R$ satisfy all the constraints, we introduce additional edits with \textsc{param} assignments based on the features of other \textit{edited} hexahedrons. 
For example, if a cabinet door is rotated about its hinge, its handle must also be rotated about the same axis. By introducing edits with \textsc{param} assignments based on the door's features, we can discover such an edit.
For simplicity, such edits are later referred to in our experiments as ~\textit{nhbd-edits}.

\noindent
\textbf{Handing UNSAT}
Despite the extensive search, sometimes the constraints in $\mathbb{C}$ may not be mutually satisfiable. 
Under such circumstances, we select the \textit{minimally constraint breaking} edit, i.e. the edit which breaks the least amount of constraints. This is done by recording for each solution in $F$, the number of constraints it breaks and selecting the one which breaks the fewest constraints. 
We simply refer to edits introduced in this way as \textit{breaking-edits}

\section{LLM Augmented Inference}
\label{sec:llm}

In the previous section, we introduced \textsc{Analytical Edit Propagation (AEP)}, which utilizes a \textit{seed-edit} inferred by the LLM to discover necessary secondary edits.
We now explore how we deploy LLMs in tandem with \textsc{AEP} to infer parameterized editing programs. First, we address a limitation of naive edit propagation, present also in traditional techniques~\cite{comp_prop, iwires}, and propose our solution to overcome it.

\subsection{A Limitation of Naive Edit Propagation}
Edit propagation relies on two key assumptions about the user's intent: (i) The user wishes to preserve inter-part relationships as much as possible, and (ii) The user prefers the simplest possible secondary modification. However, these assumptions do not always hold.

During edit propagation, we attempt to restore all inter-part geometric relationships. However, achieving certain edits may require allowing some relationships to remain broken. For example, even if the front legs of a chair are symmetric to the back legs, the user may desire to edit only the back legs in a symmetry-breaking fashion. Similarly, using the simplest possible secondary modifications, from an ARAP~\cite{ARAP_modeling} perspective, might not always match user intent. For instance, when widening a chair's seat, the simplest modification to the legs is to shift them. However, the user might instead prefer to tilt the legs (cf. Figure~\ref{fig:edit_extension}).

Performing these edits with traditional methods requires significant manual effort. Users must remove the reflection relationship between the front and back legs to support the first edit and add an attachment relationship between the legs and the floor for the second edit. This underscores a key limitation of traditional edit propagation: \textit{the structure invoked in the shape must be modified to support different edits}. Consequently, with prior methods, users have to manually adjust the shape's structure to ensure the edits generated align with their intent.

\subsection{Dynamic Structure Modification Using LLMs}

While manually performing these actions is laborious, verbally specifying them in an edit request is straightforward. As our system supports unstructured natural language input, we use LLMs to interpret these requirements from the edit request and automatically update the shape's structure to facilitate the edit. This approach eliminates the need for users to manually adjust the shape's structure, leveraging LLMs to perform these tasks instead. Based on the edit request, the LLM infers (i) the \textit{seed-edit}, (ii) the \textit{relation-validity} for each symmetry relationship in the shape, and (iii) the edit \textit{type-hints} for the different parts.

When a relation is deemed invalid, we delete the corresponding constraints. When a part has a specified \textit{type-hint}, we use it to filter the edit candidates generated during edit propagation (i.e. we filter them out from $\mathbb{E}_R$). This enables us to directly target the specific edits the user desires, unlike prior approaches that require additional constraints to achieve such complex modifications. 
With this approach, we can accommodate edit requests that conflict with symmetry relationships in the shape or require specific editing operations for secondary parts.

\subsubsection{Prompting Workflows}
We infer the three quantities with three separate prompting workflows. For each task, the LLM is provided with a verbal description of the parts in the shape, the user's edit request, and a set of task-specific instructions. The LLM returns an executable Python snippet, which is parsed and executed to receive the LLM's response. For the \textit{seed-edit} inference, the LLM is also given an API to create the edit operators. To avoid confusion between different relations, \textit{relation-validity} is inferred for each symmetry relation independently with a separate prompt. For inferring explicit \textit{type-hints}, we found the LLM to be most reliable when limited to three abstract types: `translate', `rotate', and `scale'.

We improve the quality of the LLM response by using prompting techniques such as Chain of Thought (CoT)~\cite{wei2023chainofthought}, in-context examples~\cite{in-context}, and reminders~\cite{makatura2023large}. Additionally, we perform majority-voting~\cite{voting} by aggregating results from multiple independent LLM calls and using the modal response. Voting improves our system significantly, even with only five samples (refer to Table~\ref{table:llm}). Since there is no interdependence between the prompts, we prompt the LLM for all tasks in parallel, maintaining response times comparable to a single API call.

\begin{figure*}
    \centering
    \includegraphics[width=1.0\linewidth]{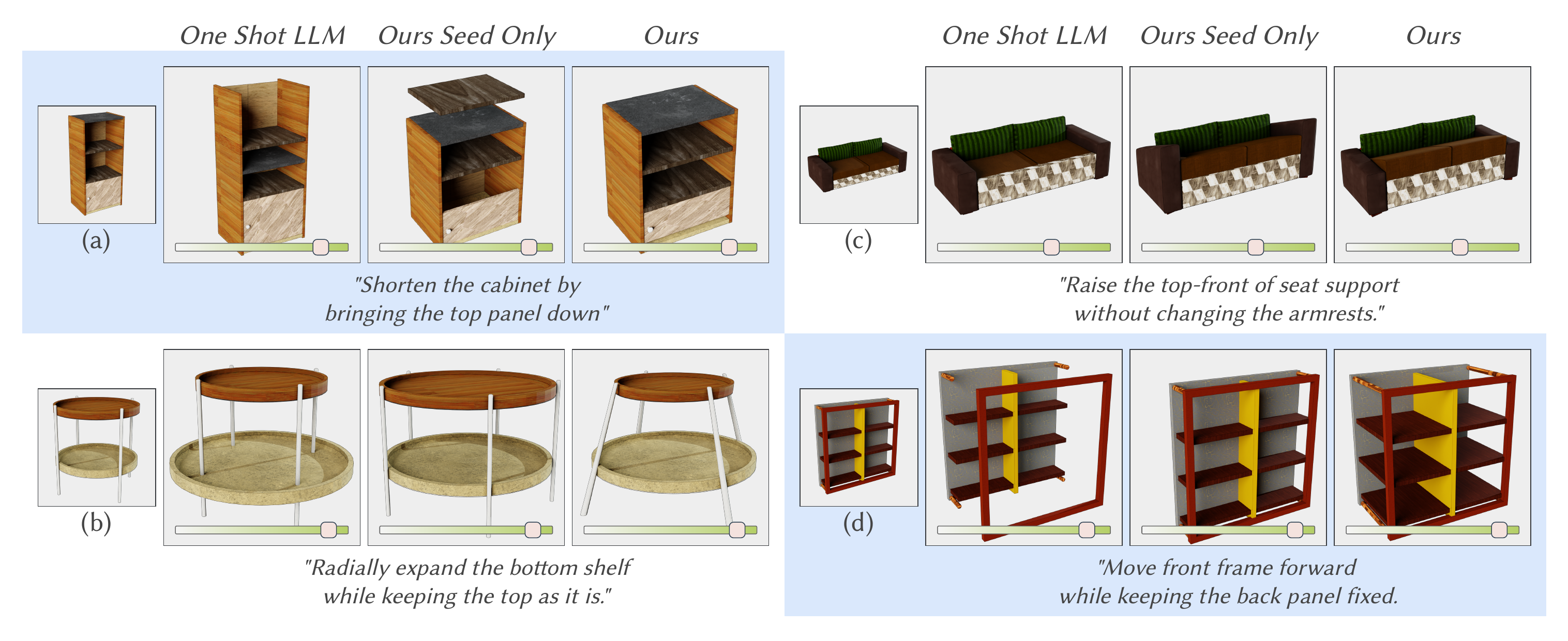}
    \caption{
    We compare parametric editing of 3D assets across three system variants. The \textit{One Shot LLM} produces unrealistic part intersections (b, c) and fails to propagate corrective edits (a, d). The \textit{Ours Seed Only} variant with \textsc{Analytical Edit Propagation (AEP)} produces consistent variations, but can fail to align with the input edit intent. 
    Our full system (\textit{Ours}), which includes edit \textit{type-hints} and \textit{relation-validity}, produces edits that closely match user requests.
    }
    \vspace{-1.25em}
    \label{fig:qualitative}
\end{figure*}

\section{Experiments}

We now present experiments which evaluate the editing programs inferred by our system. We additionally perform ablation experiments on the different components of our system to provide more insights.
Note that we utilize OpenAI's (text-only) GPT-4~\cite{openai2023gpt4} as the LLM for all of our experiments.

\subsection{Datasets}

We evaluate our method over 3D models sourced from 2 datasets. 
A \textit{dev-set} is curated using 3D mesh models from the PartNet~\cite{partnet} dataset, while the \textit{test-set} is composed of 3D models from the CoMPaT3D++ dataset~\cite{slim2023_3dcompatplus}.
The \textit{dev-set} includes $51$ part-segmented meshes sampled from five categories of man-made objects: \textit{Chair}, \textit{Table}, \textit{Couch}, \textit{Storage Furniture}, and \textit{Bed}. 
We used the \textit{dev-set} as a benchmark while developing the system, and models in this set contain from $5$ to $81$ parts (median $17$), and from $4$ to $318$ (median $40$) relations, covering simple to complex geometries. 
The \textit{test-set} contains $50$ models sourced from $21$ different categories within the CoMPaT3D++ dataset. This set is used to verify the efficacy of our system beyond the object categories present in the \textit{dev-set}. Thus, the \textit{test-set} also includes objects from uncommon categories such as \textit{Gazebos}, \textit{Bird Houses}, and \textit{Fans}.

All models are paired with manually written natural language edit requests to create \textit{(shape, edit request)} pairs used as input to our system. As we show in the qualitative examples, the edit requests encompass a wide variety of modifications.
Additionally, we annotate each pair in the \textit{dev-set} with a `ground-truth' (GT) editing program derived through a \textit{Human-Solver} inference process, where the LLM is replaced by an expert user. Note that both the GT programs and the programs inferred by each variant we consider are written in the DSL introduced in Section~\ref{sec:DSL}.

\subsection{Language-based Parameterized Editing of 3D Assets}
\label{sec:eval_infer}

First, we evaluate our system's efficacy in enabling natural-language-based parametric editing of 3D assets. 
As described earlier, our system achieves this by inferring parameterized editing programs that align with natural language edit requests. 
Notably, prior works are not well-suited for this task. 
Neural approaches~\cite{shapewalk, achlioptas2023shapetalk}, although powerful, do not support high-resolution 3D assets, and prior edit-propagation approaches~\cite{iwires, comp_prop} cannot be controlled via natural language. 
Beyond this, neither of these paradigms support \textit{parametric} editing.
Since no prior work adequately addresses this problem, we introduce alternative realizations of our system for comparison:
\begin{packed_enumerate}
\item \textit{One Shot LLM:} As detailed in Section~\ref{sec:task}, this approach involves providing the LLM with all necessary information to infer the entire editing program in a single step.
\item \textit{Ours Seed Only:} This baseline uses an LLM to infer the \textit{seed-edit} and employs \textit{Analytical Edit Propagation (AEP)} to generate the entire editing program.
\item \textit{Ours:} Our full system, which utilizes the LLM to specify the \textit{seed-edit},~\textit{relation-validity}, and edit~\textit{type-hints}. These components are then used during \textsc{AEP} to produce edits that closely align with the user's intent.
\end{packed_enumerate}

\begin{table}[t!]
    \centering
    \begin{tabular}{l c c}
        \toprule
        & Preference Rate
        \\
        \midrule
        \textit{Ours} vs. \textit{One Shot LLM} & $81.06\%$
         
        \\
        \textit{Ours} vs. \textit{Ours Seed Only} & $75.59\%$
        \\
        \bottomrule
    \end{tabular}
    \caption{\textbf{Our system is preferred}:
Results of a two-alternative forced-choice perceptual study comparing our system against two alternate realizations.
    Our system (denoted \textit{Ours}) is preferred in an overwhelming majority of the judgements.
    }
    \vspace{-1.25em}
    \label{tab:perceptual_study}
\end{table}

\subsubsection{Analysis of Human Preference}
Using the \textit{test-set}, we conducted a two-alternative forced-choice perceptual study to compare these method variants. 
We recruited $32$ participants for the task. 
Each participant was shown a series of comparisons, resulting in a total of $1295$ judgements. 
Each comparison included a natural language edit request and two videos showing a 3D shape being edited with the inferred editing programs, with the control parameter smoothly varying across its range (from $0$ to an automatically set upper bound, $\tau$).
Participants were tasked with selecting their preferred method, based on the instruction to: {``Select the method which better satisfies the input editing prompt and results in a better edited shape}''. 
Note that we elide comparisons where the programs inferred by the compared methods match ($18\%$ of the comparisons in total).

Table~\ref{tab:perceptual_study} presents the results of our experiment. Our method is preferred over \textit{One-shot LLM} in $81.1\%$ of the judgements. \textit{One-shot LLM} often fails to construct meaningful editing programs, demonstrating the failure cases discussed in Section~\ref{sec:task}. In contrast, by leveraging \textit{AEP}, our method produces cohesive editing programs that respect the inter-part relationships.
Compared to \textit{Ours Seed Only}, our method is preferred in $75.6\%$ of comparisons. Without the ability to provide type hints or disable symmetry relations, \textit{Ours Seed Only} often infers programs that respect inter-part relations but fail to align with the user's edit intent.
Note that \textit{Ours Seed Only} represents a \textit{purely} analytical variant of the prior edit propagation method~\cite{comp_prop}.
This indicates that while editing with prior \textit{analyse-and-edit} methods, users often need to manually adjust multiple settings, such as enabling/disabling relations, to perform their desired edits.
In contrast, our system allows users to simply state their edit intent in natural language, and we task the LLM with automatically adjusting these settings.
We additionally provide all the videos from the perceptual study in our supplementary materials, allowing for independent verification of our results.

\begin{table}[t!]
    \centering
    \begin{tabular}{lccc}
        \toprule
        & $\mathcal{J}(Prog) (\uparrow) $ & $\mathcal{D}(Geo) (\downarrow)$ & $\textsc{\%Rel}(\uparrow)$   \\
        \midrule
    \textit{One Shot LLM} & $0.31$ & $8.30$ & $61.71\%$\\
    \textit{Ours Seed Only} & $0.53$ & $8.62$ & $83.47\%$ \\
    \textit{Ours} & $\mathbf{0.70}$ & $\mathbf{4.01}$ & $\mathbf{91.57\%}$ \\
        \bottomrule
    \end{tabular}
    \caption{
    \textbf{Our method is closer to GT}:
    We compare the different system realizations against `GT' annotations. Our system (\textit{Ours}) obtains the best performance over metrics that measure the proximity to the `GT' annotations across \textit{Programmatic}, \textit{Geometric} and \textit{Structural} aspects.
    }
    \label{tab:gt}
    \vspace{-1.25em}
\end{table}

\subsubsection{Analysis of Inferred Programs}
Next, we compare the programs generated by these methods against the manually annotated ground-truth (GT) editing programs on the \textit{dev-set}. 
We evaluate the quality of the editing programs using three criteria:
1) \textit{Programmatic} ($\mathcal{J}(prog)$): This metric assesses how closely the inferred programs match the GT programs. 
2) \textit{Geometric} ($\mathcal{D}(geo)$): This metric measures the geometric distance between the shape edited with the inferred programs and the shape edited with the GT programs. 
3) \textit{Structural} ($\%Rel$): This metric quantifies the percentage of inter-part relations whose state (broken vs. maintained) matches the relation's state under the ground truth (GT) program.
Further details are provided in the supplementary material.

\begin{figure*}
    \centering
    \includegraphics[width=1.0\linewidth]{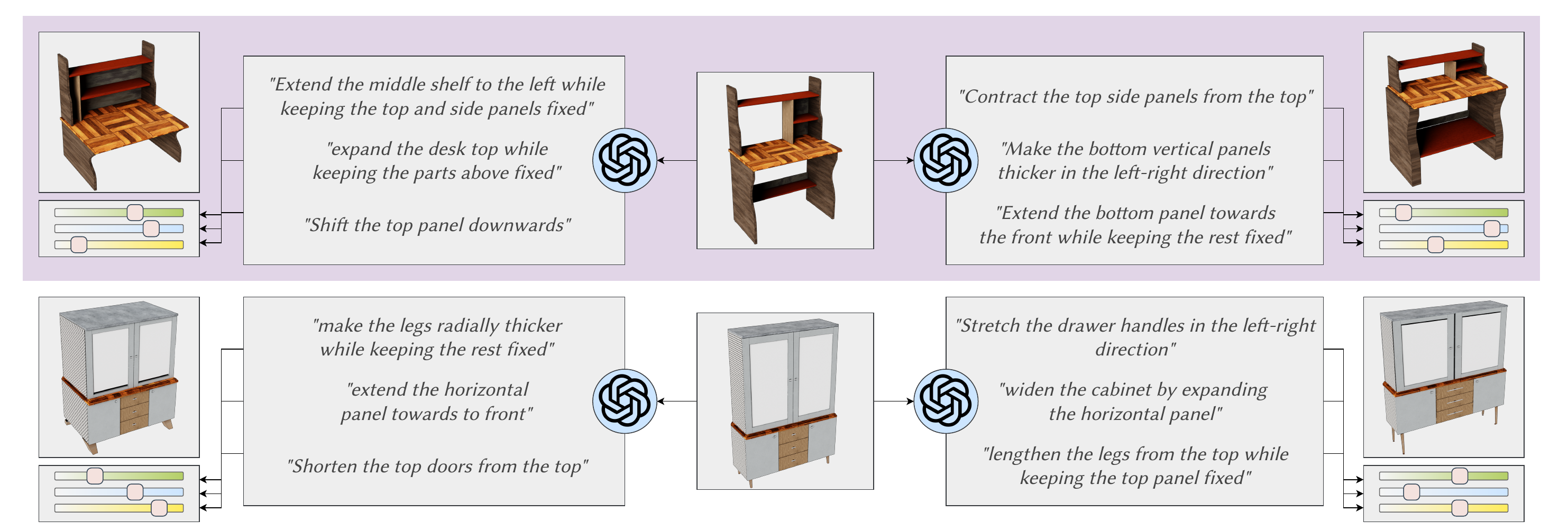}
    \caption{ 
\textbf{Proxydural Modeling}: Leveraging the open-world knowledge of LLMs, we synthesize edit requests for a given shape to enable \textit{automatic} procedural models, termed \textit{Proxydural} due to the use of bounding proxy deformations. Our system allows multiple proxydural models for the same shape, a capability not possible with prior approaches~\cite{bokeloh}.
    }
    \vspace{-1.25em}
    \label{fig:procedural}
\end{figure*}
We present the results of this experiment in Table~\ref{tab:gt}. 
First, we note that the \textit{One-shot LLM} approach fails drastically across the three criterion, which aligns with the preference rates observed in the human study. 
Secondly, we observe that omitting the \textit{type-hints} and \textit{relation-validity} steps adversely affects the \textit{Our Seed Only} approach as well. 
Specifically, \textit{Our Seed Only} results in a very high $\mathcal{D}(geo)$ measure, indicating that although \textsc{AEP} restores the inter-part relations, the edits inferred without \textit{type-hints} cause the edited shape to geometrically deviate from the intended shape. 
Finally, we highlight that our system outperforms the others across all three metrics.

We also present qualitative examples of editing programs inferred from different baselines in Figure~\ref{fig:qualitative}. The improvement in the quantitative metrics is evident in these qualitative examples. The baselines produce visible artifacts, such as intersections between parts, and often violate the intent of the requested edits.

\begin{table}[t!]

\begin{center}
\begin{tabular}{lcccc}
    \toprule
    & $Acc(SE) (\uparrow)$ & $Acc(R) (\uparrow)$ & $\mathcal{J}(T) (\uparrow)$ &  $Match (\uparrow)$ \\
    \midrule
    \textit{Ours}&  $\mathbf{76.47}$ & $\mathbf{88.91}$ & $\mathbf{73.92}$ & $\mathbf{35.29}$  \\
    \midrule
    \textit{- Voting} & $68.62$ & $\mathbf{88.91}$ & $68.02$ & $25.49$\\
    \textit{- CoT} & $72.54$ & $88.17$ & $71.69$ & $27.45$ \\
    \textit{- InContext}& $66.66$ & $87.68$ & $65.75$ & $23.52$ \\
    \midrule
    \textit{Naive} & $68.62$ & $86.99$ & $65.36$ & $19.60$ \\
    \bottomrule
\end{tabular}

\caption{\textbf{Prompting Ablation}: we report the LLM's accuracy at inferring the \textit{seed-edit} ($Acc(SE)$), \textit{relation-validity} ($Acc(R)$), and edit \textit{type-hints} ($\mathcal{J}(T)$), along with the fraction of inputs where the LLM infers everything correctly ($Match$). Voting and in-context examples significantly enhance accuracy, while the naive approach shows a marked decrease.}
    \vspace{-1.25em}

\label{table:llm}

\end{center}
\end{table}

\begin{table}[t!]
    \centering
    \begin{tabular}{lccc}
        \toprule
        & $\mathcal{J}(Prog) (\uparrow) $ & $\mathcal{D}(Geo) (\downarrow)$ & $\textsc{\%Rel}(\uparrow)$   \\
        \midrule
        \textit{Ours} &  $\mathbf{0.70}$ & 4.01 & $\mathbf{91.57\%}$ \\
        \midrule
         \textit{- nhbd} & $\mathbf{0.70}$ & 4.21 & 91.35\%  \\
        \textit{- breaking}& 0.68 & \textbf{3.46 }& 86.41\%  \\
        \midrule
        \textit{Naive} &0.67 & 4.31 & 85.4\% \\
        \bottomrule
    \end{tabular}
    \caption{ \textbf{Quality of programs inferred by the Solver}: Removing \textit{nhbd-edits} results in higher geometric distance ($\mathcal{D}(Geo)$), while removing \textit{breaking-edits} leads to more structural distance ($\textsc{\%Rel}$). The naive approach that removes both of these options is the least effective.
    }
    
    \vspace{-1.25em}
    \label{tab:solver}
\end{table}

\subsection{Analysing the System Design}
\label{sec:ablation}

In this section, we present an ablative analysis of the LLM prompting workflow and the \textsc{AEP} solver to elucidate the impact of various design decisions.

\subsubsection{Analysis of the Prompting Workflow}
Our system employs separate prompts to infer the \textit{seed edit}, \textit{relation validity}, and edit \textit{type hints}. 
We measure the LLM's accuracy at inferring these three terms by comparing its output to that of a human annotator. 
Additionally, we perform a subtractive ablation by individually removing majority voting~\cite{voting}, Chain-of-Thought (CoT)~\cite{wei2023chainofthought}, and in-context examples~\cite{in-context} to measure their impact on the LLM's accuracy.
We report four metrics: 1) $Acc(SE)$, for seed edit accuracy; 2) $Acc(R)$, for relation validity accuracy; 3) $\mathcal{J}(T)$, for type hint accuracy; and 4) $Match$, for the fraction of input pairs where all LLM-inferred quantities match human annotations.

The results are presented in Table~\ref{table:llm}. 
As demonstrated, there is a significant gap in accuracy between our prompting workflow and the \textit{Naive} approach, which does not include voting, chain-of-thought (CoT) reasoning, or in-context examples.
Among these techniques, voting and in-context examples have a more pronounced effect on performance, whereas the absence of CoT results in a relatively smaller decline in accuracy. 
Although not directly comparable, we find that all models struggle most with correctly setting the seed edit and the type hints. 
This difficulty likely arises from the need to construct the seed edit using a complex API and the necessity of inferring type hints for many parts of the shape. 
Finally, this experiment underscores the task and dataset complexity, as even our best approach fully matches the human annotation for only approximately 35\% of the input pairs.
Note that due to redundancies in the shape structure and the presence of multiple ways to satisfy an edit request, effective editing programs can be inferred even when not all inferred quantities match the ground truth.
Consequently, annotator evaluations indicate that programs produced with our method (\textit{Ours}) matched the intended edits in 72.5\% of the inputs. 

\subsubsection{Analysis of the AEP Solver}
Next, we perform an ablative analysis of our \textit{AEP} solver, focusing on the features introduced in Section~\ref{sec:aep_ext}: (i) \textit{nhbd-edits}, which creates edit candidates with parameters specified with features of other edited parts, and (ii) \textit{breaking-edits}, which use the \textit{minimally constraint-breaking} edit when no edit satisfies all constraints. We compare the inferred programs to the ground truth (GT) programs using the metrics from Section~\ref{sec:eval_infer}, and provide results in Table~\ref{tab:solver}. Removing \textit{nhbd-edits} increases \textit{geometric} distance, while removing \textit{breaking-edits} increases \textit{structural} distance. This aligns with our intuition: \textit{nhbd-edits} improve parameterization, affecting geometric distance, while \textit{breaking-edits} reduce the number of broken relations, affecting structural distance when elided.

\begin{figure}
    \centering
    \includegraphics[width=1.0\linewidth]{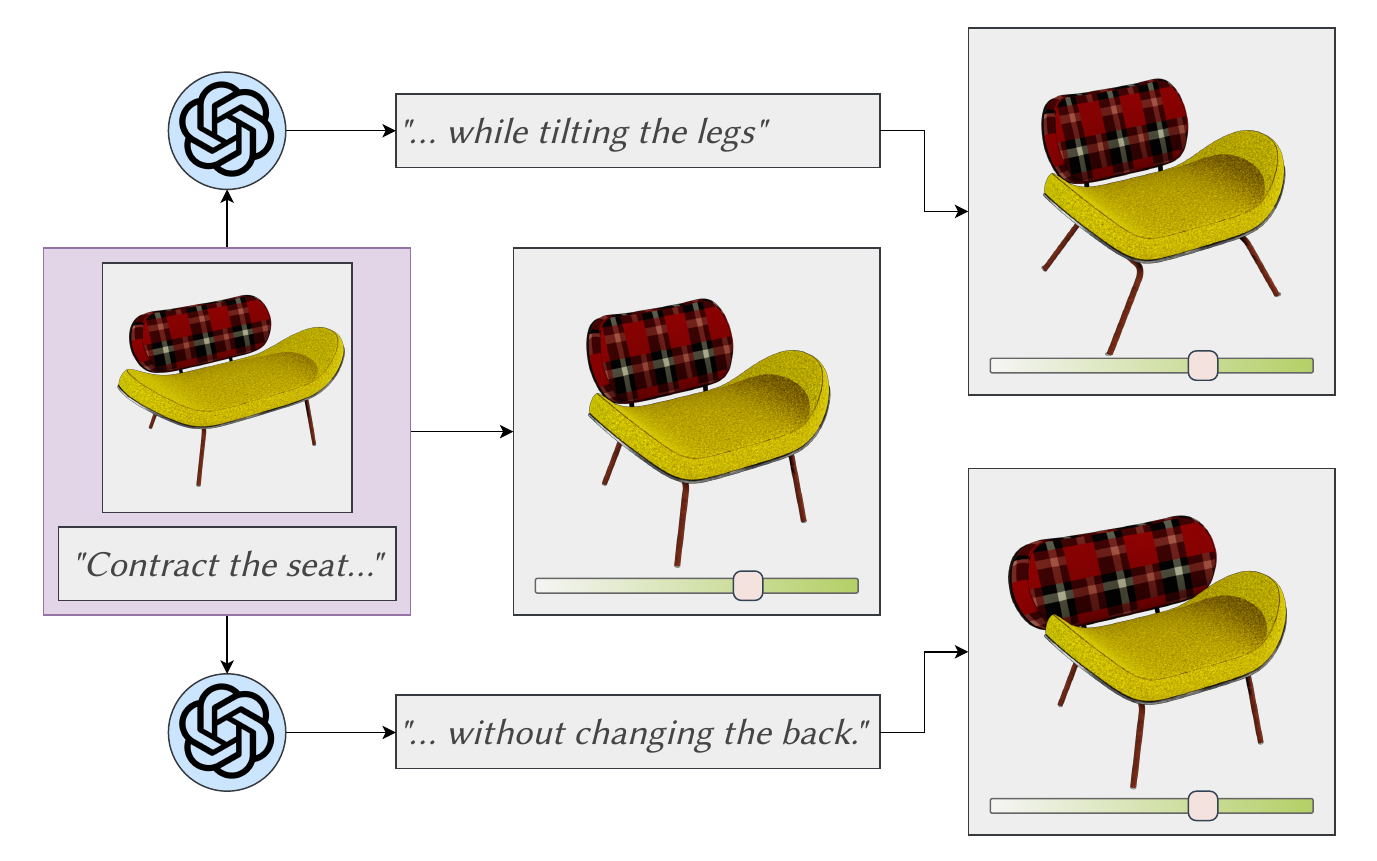}
    \caption{
    \textbf{Shape Variation Family}: Given a 3D asset and an under-specified edit request, we leverage LLMs to synthesize potential extensions of the request. By inferring programs for both the extended and original requests, we create a set of related but distinct parametric models, each adhering to the initial edit request while supporting different shape variations. 
    }
    \vspace{-1.25em}
    \label{fig:edit_extension}
\end{figure}

\section{Applications}

Our editing system acts as a translator, converting natural language prompts into editing programs. With this system in hand, we can further explore how to employ the creative capabilities of LLMs to enable tantalizing new applications. 
We demonstrate two such applications: \textit{Automatic Proxydural Modeling} and Generating \textit{Shape Variation Families}.
Note that we utilize the OpenAI's (Vision) GPT-4~\cite{openai2024gpt4} as the LLM for these applications, and provide the prompts utilized for all applications in the supplementary matarials.

\subsection{Automatic Proxydural Modeling}
\label{sec:proxydural}
Our editing programs incorporate analytical edits, representing modified parts as parameterized functions of control parameters exposed to the user. 
By leveraging function composition, these programs can be automatically stacked, allowing a part edited by one program to serve as the input for another. 
This capability enables the creation of (approximately) procedural models of any given shape through the following steps: (i) gather multiple independent editing requests, (ii) infer an editing program for each request, each with a single independent control parameter, and (iii) stack these edits using function composition. 
This approach allows users to explore shape variations through a set of sliders, akin to interacting with procedural models. Since the shapes are edited using their bounding proxies, we term this approach \textit{Proxydural Modeling}.

In cases where users wish to automatically explore variations of a given shape, our system can still be effectively utilized. The key is to identify interesting modes of variation and automatically craft corresponding editing requests. Once these editing requests are defined, our system can use them to generate the \textit{Proxydural Model}.
To achieve this, we leverage the impressive world knowledge and natural language capabilities of LLMs. Given a segmented shape, we prompt the LLM to generate edit requests that capture interesting variations. By providing in-context examples and task-specific instructions, we ensure that the LLM produces requests compatible with our system. These requests are then used to automatically create the \textsc{Proxydural Model}.
In Figure~\ref{fig:procedural}, we present multiple \textsc{Proxydural Models} automatically crafted by the LLM. Unlike prior inverse procedural modeling approaches~\cite{bokeloh}, our system enables the creation of multiple \textit{Proxydural Models} for a single input geometry, each facilitating the exploration of different shape variations. 

\subsection{Creating a Shape Variation Family}

Natural language edit requests can often be under-specific, particularly when it comes to detailing secondary edits. Consequently, edit requests typically support multiple possible realizations. The \textit{minimal-deformation}, \textit{maximally-relation-preserving} edits produced by edit propagation approaches represent only a single interpretation of the request. However, different users may prefer different secondary edits. Therefore, producing editing programs that each model a different interpretation of the request is desirable. We refer to these sets of related but distinct editing programs as \textit{Shape Variation Families}.

To enable this exploration, we create multiple variations of the initial editing request, retaining the user-specified primary edit while exploring different secondary effects. These varied requests are then used to generate distinct editing programs, each resulting in a unique shape variation.
We leverage the general world knowledge about object categories contained in LLMs to accomplish this task. Given the user's primary edit request, we prompt the LLM to generate variations of the request, specifically targeting different secondary effects. As with the previous application, we provide in-context examples and task-specific instructions to ensure that the LLM produces requests compatible with our system.
Once the editing programs are generated for all the varied prompts, they are presented to the user. The user can then explore different interpretations of the edit request, and select the one most aligned with their intent. In Figure~\ref{fig:edit_extension} we show an example of a \textit{Shape Variation Family} generated with this approach.

\section{Conclusions}

We have introduced \textsc{ParSEL: Parameterized Shape Editing with Language}, a novel shape editing system that combines the intuitiveness of natural language with the precision of parametric control to perform geometric edits on 3D assets. 
Given semantically labeled 3D meshes and a natural language edit request, our system provides adjustable parameters to control how the shape is edited. 
This capability is achieved through \textit{parameterized editing programs}, inferred by a neuro-symbolic module that combines LLM prompting with CAS solvers. 
Central to this module is \textsc{Analytical Edit Propagation (AEP)}, a novel technique that extends editing programs by considering inter-part geometric relationships. 
Our experiments demonstrated that our system infers editing programs that are more closely aligned with user intent compared to other baselines.
We also showcased exciting applications made possible by our system, such as automatic conversion of a high-quality 3D asset into an approximately procedural model. 

\subsection{Limitations and future work}

While \textsc{ParSEL} is the first system capable of supporting controllable shape edits from natural language, it does have a few limitations:

\textit{(i) Search Efficiency}: A downside of our approach is the computational expense of the analytical edit search. While the median time it takes to search for editing programs is approximately 30 seconds, in certain conditions it can take beyond 500 seconds for very complex shapes. This is due to our simple and generous edit sampling process, where the solver may evaluate an extensive array of potential edit candidates under certain conditions. This could likely be mitigated to a large extent with the addition of a `smarter' edit sampling strategy.
For instance, analyzing the parametric form of attachment points can help identify suitable editing operations and prune the search space. 
Alternatively, employing an early exit model, where search stops once a satisfying edit is found, may also improve the run-time of the system (though integration with parallelization may be non-trivial).

\begin{figure}
    \centering
    \includegraphics[width=1.0\linewidth]{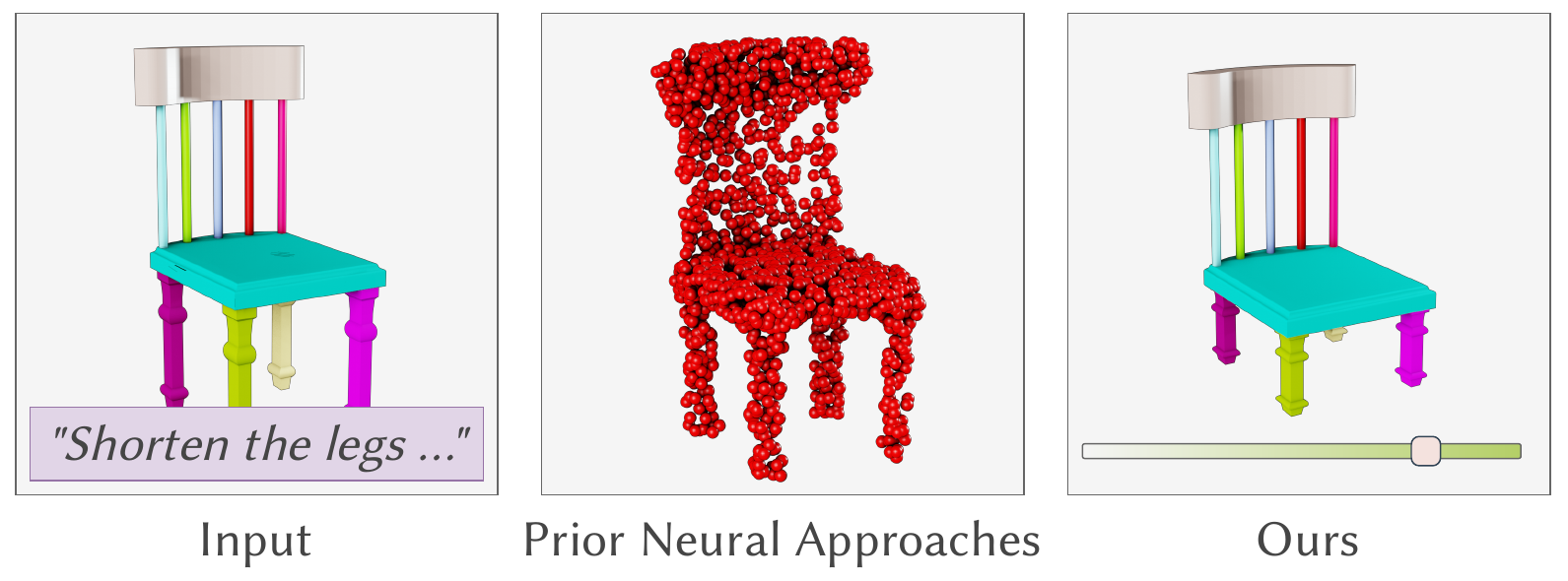}
    \caption{
    While prior language-based 3D shape editing methods~\cite{achlioptas2023shapetalk} support a broader range of operations on unsegmented shapes, \textsc{ParSEL} delivers \textit{controllable}, \textit{disentangled} editing of high-quality textured 3D assets.
    Each approach is tailored to suit different needs.
    }
    \vspace{-1.25em}
    \label{fig:shapetalk}
\end{figure}

\textit{(ii) Limited Edits}: Our system currently supports a limited range of part deformations - only those expressible as affine transforms of the bounding proxy or its features. 
We contrast this with the prior neural approaches~\cite{achlioptas2023shapetalk} in Figure~\ref{fig:shapetalk}, which despite other drawbacks, support a wider range of shape edits.
To accommodate a wider range of edit requests, our system needs to support additional deformation functions, such as making parts spherical or cylindrical. 
A key challenge in supporting novel deformations is the ability to analytically specify the state of inter-part relations under these deformation functions. 
This makes iterative non-analytic deformations functions, such as ARAP energy based deformation~\cite{ARAP_modeling}, incompatible with our system.

\textit{(iii) Limited Shape Structures}: While our shape representation is effective for many man-made 3D objects, it has its limitations. Bounding boxes may not be the best control cage for all parts and other control cages may be required. 
Although our system can support arbitrary control cages, as we use harmonic coordinates to parameterize the attachment relations, their effect on the solver's success is unclear. 
Additionally, we currently model a limited set of symmetry relations. Two common forms observed in 3D assets that we identify are (i) parts formed by joining points on two different curves, often seen in the design of back supports in chairs and beds, and (ii) wallpaper symmetry groups, which involve translation symmetry along two axes, commonly seen in building facades. Extending our system to support these additional structures would greatly enhance its applicability.

We believe that our system makes 3D asset editing more controllable and intuitive for artists. 
With the rise of text-to-3D approaches, creating new models has become easier, yet editing them remains a challenge. 
By integrating our system with auto-segmentation techniques tools~\cite{zhou2023partslip}, we can enable intuitive and precise editing of models generated from text-to-3D approaches. 
Finally we emphasize that our system was most effective with careful and directed LLM integration; we task the LLM with performing only high-level translation tasks, and rely on symbolic solvers to perform the necessary geometric analysis. 
Looking ahead, we hope this work provides a potent framework for a reliable integration of LLMs with symbolic reasoning to support shape analysis.

\clearpage

\bibliographystyle{ACM-Reference-Format}
\bibliography{main.bib}


\begin{thebibliography}{56}


\ifx \showCODEN    \undefined \def \showCODEN     #1{\unskip}     \fi
\ifx \showDOI      \undefined \def \showDOI       #1{#1}\fi
\ifx \showISBNx    \undefined \def \showISBNx     #1{\unskip}     \fi
\ifx \showISBNxiii \undefined \def \showISBNxiii  #1{\unskip}     \fi
\ifx \showISSN     \undefined \def \showISSN      #1{\unskip}     \fi
\ifx \showLCCN     \undefined \def \showLCCN      #1{\unskip}     \fi
\ifx \shownote     \undefined \def \shownote      #1{#1}          \fi
\ifx \showarticletitle \undefined \def \showarticletitle #1{#1}   \fi
\ifx \showURL      \undefined \def \showURL       {\relax}        \fi
\providecommand\bibfield[2]{#2}
\providecommand\bibinfo[2]{#2}
\providecommand\natexlab[1]{#1}
\providecommand\showeprint[2][]{arXiv:#2}

\bibitem[Achlioptas et~al\mbox{.}(2023)]%
        {achlioptas2023shapetalk}
\bibfield{author}{\bibinfo{person}{Panos Achlioptas}, \bibinfo{person}{Ian Huang}, \bibinfo{person}{Minhyuk Sung}, \bibinfo{person}{Sergey Tulyakov}, {and} \bibinfo{person}{Leonidas Guibas}.} \bibinfo{year}{2023}\natexlab{}.
\newblock \showarticletitle{ShapeTalk: A Language Dataset and Framework for 3D Shape Edits and Deformations}. In \bibinfo{booktitle}{\emph{Proceedings of the IEEE/CVF Conference on Computer Vision and Pattern Recognition (CVPR)}}. \bibinfo{pages}{12685--12694}.
\newblock


\bibitem[Aliaga et~al\mbox{.}(2007)]%
        {aliaga2007style}
\bibfield{author}{\bibinfo{person}{Daniel~G Aliaga}, \bibinfo{person}{Paul~A Rosen}, {and} \bibinfo{person}{Daniel~R Bekins}.} \bibinfo{year}{2007}\natexlab{}.
\newblock \showarticletitle{Style grammars for interactive visualization of architecture}.
\newblock \bibinfo{journal}{\emph{IEEE transactions on visualization and computer graphics}} \bibinfo{volume}{13}, \bibinfo{number}{4} (\bibinfo{year}{2007}), \bibinfo{pages}{786--797}.
\newblock
\urldef\tempurl%
\url{https://doi.org/10.1109/TVCG.2007.1024}
\showDOI{\tempurl}


\bibitem[Bernstein and Li(2015)]%
        {lillicon}
\bibfield{author}{\bibinfo{person}{Gilbert~Louis Bernstein} {and} \bibinfo{person}{Wilmot Li}.} \bibinfo{year}{2015}\natexlab{}.
\newblock \showarticletitle{Lillicon: using transient widgets to create scale variations of icons}.
\newblock \bibinfo{journal}{\emph{ACM Trans. Graph.}} \bibinfo{volume}{34}, \bibinfo{number}{4}, Article \bibinfo{articleno}{144} (\bibinfo{date}{jul} \bibinfo{year}{2015}), \bibinfo{numpages}{11}~pages.
\newblock
\showISSN{0730-0301}
\urldef\tempurl%
\url{https://doi.org/10.1145/2766980}
\showDOI{\tempurl}


\bibitem[Bokeloh et~al\mbox{.}(2011)]%
        {pattern_docker}
\bibfield{author}{\bibinfo{person}{Martin Bokeloh}, \bibinfo{person}{Michael Wand}, \bibinfo{person}{Vladlen Koltun}, {and} \bibinfo{person}{Hans-Peter Seidel}.} \bibinfo{year}{2011}\natexlab{}.
\newblock \showarticletitle{Pattern-aware shape deformation using sliding dockers}.
\newblock \bibinfo{journal}{\emph{ACM Trans. Graph.}} \bibinfo{volume}{30}, \bibinfo{number}{6} (\bibinfo{date}{dec} \bibinfo{year}{2011}), \bibinfo{pages}{1–10}.
\newblock
\showISSN{0730-0301}
\urldef\tempurl%
\url{https://doi.org/10.1145/2070781.2024157}
\showDOI{\tempurl}


\bibitem[Bokeloh et~al\mbox{.}(2012)]%
        {bokeloh}
\bibfield{author}{\bibinfo{person}{Martin Bokeloh}, \bibinfo{person}{Michael Wand}, \bibinfo{person}{Hans-Peter Seidel}, {and} \bibinfo{person}{Vladlen Koltun}.} \bibinfo{year}{2012}\natexlab{}.
\newblock \showarticletitle{An algebraic model for parameterized shape editing}.
\newblock  \bibinfo{volume}{31}, \bibinfo{number}{4}, Article \bibinfo{articleno}{78} (\bibinfo{date}{jul} \bibinfo{year}{2012}), \bibinfo{numpages}{10}~pages.
\newblock
\showISSN{0730-0301}
\urldef\tempurl%
\url{https://doi.org/10.1145/2185520.2185574}
\showDOI{\tempurl}


\bibitem[Brooks et~al\mbox{.}(2022)]%
        {brooks2022instructpix2pix}
\bibfield{author}{\bibinfo{person}{Tim Brooks}, \bibinfo{person}{Aleksander Holynski}, {and} \bibinfo{person}{Alexei~A Efros}.} \bibinfo{year}{2022}\natexlab{}.
\newblock \showarticletitle{InstructPix2Pix: Learning to Follow Image Editing Instructions}.
\newblock \bibinfo{journal}{\emph{arXiv preprint arXiv:2211.09800}} (\bibinfo{year}{2022}).
\newblock


\bibitem[Brown et~al\mbox{.}(2020)]%
        {in-context}
\bibfield{author}{\bibinfo{person}{Tom Brown}, \bibinfo{person}{Benjamin Mann}, \bibinfo{person}{Nick Ryder}, \bibinfo{person}{Melanie Subbiah}, \bibinfo{person}{Jared~D Kaplan}, \bibinfo{person}{Prafulla Dhariwal}, \bibinfo{person}{Arvind Neelakantan}, \bibinfo{person}{Pranav Shyam}, \bibinfo{person}{Girish Sastry}, \bibinfo{person}{Amanda Askell}, \bibinfo{person}{Sandhini Agarwal}, \bibinfo{person}{Ariel Herbert-Voss}, \bibinfo{person}{Gretchen Krueger}, \bibinfo{person}{Tom Henighan}, \bibinfo{person}{Rewon Child}, \bibinfo{person}{Aditya Ramesh}, \bibinfo{person}{Daniel Ziegler}, \bibinfo{person}{Jeffrey Wu}, \bibinfo{person}{Clemens Winter}, \bibinfo{person}{Chris Hesse}, \bibinfo{person}{Mark Chen}, \bibinfo{person}{Eric Sigler}, \bibinfo{person}{Mateusz Litwin}, \bibinfo{person}{Scott Gray}, \bibinfo{person}{Benjamin Chess}, \bibinfo{person}{Jack Clark}, \bibinfo{person}{Christopher Berner}, \bibinfo{person}{Sam McCandlish}, \bibinfo{person}{Alec Radford}, \bibinfo{person}{Ilya Sutskever}, {and}
  \bibinfo{person}{Dario Amodei}.} \bibinfo{year}{2020}\natexlab{}.
\newblock \showarticletitle{Language Models are Few-Shot Learners}. In \bibinfo{booktitle}{\emph{Advances in Neural Information Processing Systems}}, \bibfield{editor}{\bibinfo{person}{H.~Larochelle}, \bibinfo{person}{M.~Ranzato}, \bibinfo{person}{R.~Hadsell}, \bibinfo{person}{M.F. Balcan}, {and} \bibinfo{person}{H.~Lin}} (Eds.), Vol.~\bibinfo{volume}{33}. \bibinfo{publisher}{Curran Associates, Inc.}, \bibinfo{pages}{1877--1901}.
\newblock
\urldef\tempurl%
\url{https://proceedings.neurips.cc/paper_files/paper/2020/file/1457c0d6bfcb4967418bfb8ac142f64a-Paper.pdf}
\showURL{%
\tempurl}


\bibitem[Cabral et~al\mbox{.}(2009)]%
        {architecture}
\bibfield{author}{\bibinfo{person}{Marcio Cabral}, \bibinfo{person}{Sylvain Lefebvre}, \bibinfo{person}{Carsten Dachsbacher}, {and} \bibinfo{person}{George Drettakis}.} \bibinfo{year}{2009}\natexlab{}.
\newblock \showarticletitle{Structure Preserving Reshape for Textured Architectural Scenes}.
\newblock \bibinfo{journal}{\emph{Computer Graphics Forum (Proceedings of the Eurographics conference)}} (\bibinfo{year}{2009}).
\newblock
\urldef\tempurl%
\url{http://www-sop.inria.fr/reves/Basilic/2009/CLDD09}
\showURL{%
\tempurl}


\bibitem[Cascaval et~al\mbox{.}(2022)]%
        {diff_cad}
\bibfield{author}{\bibinfo{person}{D. Cascaval}, \bibinfo{person}{M. Shalah}, \bibinfo{person}{P. Quinn}, \bibinfo{person}{R. Bodik}, \bibinfo{person}{M. Agrawala}, {and} \bibinfo{person}{A. Schulz}.} \bibinfo{year}{2022}\natexlab{}.
\newblock \showarticletitle{Differentiable 3D CAD Programs for Bidirectional Editing}.
\newblock \bibinfo{journal}{\emph{Computer Graphics Forum}} \bibinfo{volume}{41}, \bibinfo{number}{2} (\bibinfo{year}{2022}), \bibinfo{pages}{309--323}.
\newblock
\urldef\tempurl%
\url{https://doi.org/10.1111/cgf.14476}
\showDOI{\tempurl}
\showeprint{https://onlinelibrary.wiley.com/doi/pdf/10.1111/cgf.14476}


\bibitem[Cheng et~al\mbox{.}(2024)]%
        {cheng2024learning}
\bibfield{author}{\bibinfo{person}{Ta-Ying Cheng}, \bibinfo{person}{Matheus Gadelha}, \bibinfo{person}{Thibault Groueix}, \bibinfo{person}{Matthew Fisher}, \bibinfo{person}{Radomir Mech}, \bibinfo{person}{Andrew Markham}, {and} \bibinfo{person}{Niki Trigoni}.} \bibinfo{year}{2024}\natexlab{}.
\newblock \showarticletitle{Learning Continuous 3D Words for Text-to-Image Generation}.
\newblock \bibinfo{journal}{\emph{arXiv preprint arXiv:2402.08654}} (\bibinfo{year}{2024}).
\newblock


\bibitem[Coquillart(1990)]%
        {Coquillart}
\bibfield{author}{\bibinfo{person}{Sabine Coquillart}.} \bibinfo{year}{1990}\natexlab{}.
\newblock \showarticletitle{Extended free-form deformation: a sculpturing tool for 3D geometric modeling}.
\newblock \bibinfo{journal}{\emph{SIGGRAPH Comput. Graph.}} \bibinfo{volume}{24}, \bibinfo{number}{4} (\bibinfo{date}{sep} \bibinfo{year}{1990}), \bibinfo{pages}{187–196}.
\newblock
\showISSN{0097-8930}
\urldef\tempurl%
\url{https://doi.org/10.1145/97880.97900}
\showDOI{\tempurl}


\bibitem[Feng et~al\mbox{.}(2023)]%
        {feng2023layoutgpt}
\bibfield{author}{\bibinfo{person}{Weixi Feng}, \bibinfo{person}{Wanrong Zhu}, \bibinfo{person}{Tsu-Jui Fu}, \bibinfo{person}{Varun Jampani}, \bibinfo{person}{Arjun~Reddy Akula}, \bibinfo{person}{Xuehai He}, \bibinfo{person}{S Basu}, \bibinfo{person}{Xin~Eric Wang}, {and} \bibinfo{person}{William~Yang Wang}.} \bibinfo{year}{2023}\natexlab{}.
\newblock \showarticletitle{Layout{GPT}: Compositional Visual Planning and Generation with Large Language Models}. In \bibinfo{booktitle}{\emph{Thirty-seventh Conference on Neural Information Processing Systems}}.
\newblock
\urldef\tempurl%
\url{https://openreview.net/forum?id=Xu8aG5Q8M3}
\showURL{%
\tempurl}


\bibitem[Gal et~al\mbox{.}(2009)]%
        {iwires}
\bibfield{author}{\bibinfo{person}{Ran Gal}, \bibinfo{person}{Olga Sorkine}, \bibinfo{person}{Niloy~J. Mitra}, {and} \bibinfo{person}{Daniel Cohen-Or}.} \bibinfo{year}{2009}\natexlab{}.
\newblock \showarticletitle{iWIRES: an analyze-and-edit approach to shape manipulation} \emph{(\bibinfo{series}{SIGGRAPH '09})}. \bibinfo{publisher}{Association for Computing Machinery}, \bibinfo{address}{New York, NY, USA}, Article \bibinfo{articleno}{33}, \bibinfo{numpages}{10}~pages.
\newblock
\showISBNx{9781605587264}
\urldef\tempurl%
\url{https://doi.org/10.1145/1576246.1531339}
\showDOI{\tempurl}


\bibitem[Gao et~al\mbox{.}(2023)]%
        {Gao_2023_SIGGRAPH}
\bibfield{author}{\bibinfo{person}{William Gao}, \bibinfo{person}{Noam Aigerman}, \bibinfo{person}{Groueix Thibault}, \bibinfo{person}{Vladimir Kim}, {and} \bibinfo{person}{Rana Hanocka}.} \bibinfo{year}{2023}\natexlab{}.
\newblock \showarticletitle{TextDeformer: Geometry Manipulation using Text Guidance}. In \bibinfo{booktitle}{\emph{ACM Transactions on Graphics (SIGGRAPH)}}.
\newblock


\bibitem[GPT-4(2023)]%
        {openai2023gpt4}
\bibfield{author}{\bibinfo{person}{OpenAI GPT-4}.} \bibinfo{year}{2023}\natexlab{}.
\newblock \bibinfo{title}{GPT-4 Technical Report}.
\newblock
\newblock
\showeprint[arxiv]{2303.08774}~[cs.CL]


\bibitem[Guerrero et~al\mbox{.}(2016)]%
        {GuerreroEtAl:PATEX:2016}
\bibfield{author}{\bibinfo{person}{Paul Guerrero}, \bibinfo{person}{Gilbert Bernstein}, \bibinfo{person}{Wilmot Li}, {and} \bibinfo{person}{Niloy~J. Mitra}.} \bibinfo{year}{2016}\natexlab{}.
\newblock \showarticletitle{{PATEX}: Exploring Pattern Variations}.
\newblock \bibinfo{journal}{\emph{ACM Trans. Graph.}} \bibinfo{volume}{35}, \bibinfo{number}{4} (\bibinfo{year}{2016}), \bibinfo{pages}{48:1--48:13}.
\newblock
\showISSN{0730-0301}
\urldef\tempurl%
\url{https://doi.org/10.1145/2897824.2925950}
\showDOI{\tempurl}


\bibitem[Guerrero-Viu et~al\mbox{.}(2024)]%
        {guerrero2024texsliders}
\bibfield{author}{\bibinfo{person}{Julia Guerrero-Viu}, \bibinfo{person}{Milos Hasan}, \bibinfo{person}{Arthur Roullier}, \bibinfo{person}{Midhun Harikumar}, \bibinfo{person}{Yiwei Hu}, \bibinfo{person}{Paul Guerrero}, \bibinfo{person}{Diego Gutierrez}, \bibinfo{person}{Belen Masia}, {and} \bibinfo{person}{Valentin Deschaintre}.} \bibinfo{year}{2024}\natexlab{}.
\newblock \showarticletitle{TexSliders: Diffusion-Based Texture Editing in CLIP Space}.
\newblock \bibinfo{journal}{\emph{arXiv preprint arXiv:2405.00672}} (\bibinfo{year}{2024}).
\newblock


\bibitem[Gupta and Kembhavi(2023)]%
        {visprog}
\bibfield{author}{\bibinfo{person}{Tanmay Gupta} {and} \bibinfo{person}{Aniruddha Kembhavi}.} \bibinfo{year}{2023}\natexlab{}.
\newblock \showarticletitle{Visual Programming: Compositional Visual Reasoning Without Training}. In \bibinfo{booktitle}{\emph{Proceedings of the IEEE/CVF Conference on Computer Vision and Pattern Recognition (CVPR)}}. \bibinfo{pages}{14953--14962}.
\newblock


\bibitem[Hu et~al\mbox{.}(2024)]%
        {hu2024scenecraft}
\bibfield{author}{\bibinfo{person}{Ziniu Hu}, \bibinfo{person}{Ahmet Iscen}, \bibinfo{person}{Aashi Jain}, \bibinfo{person}{Thomas Kipf}, \bibinfo{person}{Yisong Yue}, \bibinfo{person}{David~A. Ross}, \bibinfo{person}{Cordelia Schmid}, {and} \bibinfo{person}{Alireza Fathi}.} \bibinfo{year}{2024}\natexlab{}.
\newblock \bibinfo{title}{SceneCraft: An LLM Agent for Synthesizing 3D Scene as Blender Code}.
\newblock
\newblock
\showeprint[arxiv]{2403.01248}~[cs.CV]


\bibitem[Huang et~al\mbox{.}(2023)]%
        {aladdin}
\bibfield{author}{\bibinfo{person}{Ian Huang}, \bibinfo{person}{Vrishab Krishna}, \bibinfo{person}{Omoruyi Atekha}, {and} \bibinfo{person}{Leonidas Guibas}.} \bibinfo{year}{2023}\natexlab{}.
\newblock \showarticletitle{Aladdin: Zero-Shot Hallucination of Stylized 3D Assets from Abstract Scene Descriptions}.
\newblock \bibinfo{journal}{\emph{arXiv preprint arXiv:2306.06212}} (\bibinfo{year}{2023}).
\newblock


\bibitem[Huang et~al\mbox{.}(2024)]%
        {huang2024blenderalchemy}
\bibfield{author}{\bibinfo{person}{Ian Huang}, \bibinfo{person}{Guandao Yang}, {and} \bibinfo{person}{Leonidas Guibas}.} \bibinfo{year}{2024}\natexlab{}.
\newblock \bibinfo{title}{BlenderAlchemy: Editing 3D Graphics with Vision-Language Models}.
\newblock
\newblock
\showeprint[arxiv]{2404.17672}~[cs.CV]


\bibitem[Jones et~al\mbox{.}(2021)]%
        {jones2021shapeMOD}
\bibfield{author}{\bibinfo{person}{R.~Kenny Jones}, \bibinfo{person}{David Charatan}, \bibinfo{person}{Paul Guerrero}, \bibinfo{person}{Niloy~J. Mitra}, {and} \bibinfo{person}{Daniel Ritchie}.} \bibinfo{year}{2021}\natexlab{}.
\newblock \showarticletitle{ShapeMOD: Macro Operation Discovery for 3D Shape Programs}.
\newblock \bibinfo{journal}{\emph{ACM Transactions on Graphics (TOG), Siggraph 2021}} (\bibinfo{year}{2021}).
\newblock


\bibitem[Jones et~al\mbox{.}(2023)]%
        {jones2023ShapeCoder}
\bibfield{author}{\bibinfo{person}{R.~Kenny Jones}, \bibinfo{person}{Paul Guerrero}, \bibinfo{person}{Niloy~J. Mitra}, {and} \bibinfo{person}{Daniel Ritchie}.} \bibinfo{year}{2023}\natexlab{}.
\newblock \showarticletitle{ShapeCoder: Discovering Abstractions for Visual Programs from Unstructured Primitives}.
\newblock \bibinfo{journal}{\emph{ACM Transactions on Graphics (TOG), Siggraph 2023}} \bibinfo{volume}{42}, \bibinfo{number}{4}, Article \bibinfo{articleno}{49} (\bibinfo{year}{2023}).
\newblock


\bibitem[Joshi et~al\mbox{.}(2007)]%
        {harm_coords}
\bibfield{author}{\bibinfo{person}{Pushkar Joshi}, \bibinfo{person}{Mark Meyer}, \bibinfo{person}{Tony DeRose}, \bibinfo{person}{Brian Green}, {and} \bibinfo{person}{Tom Sanocki}.} \bibinfo{year}{2007}\natexlab{}.
\newblock \showarticletitle{Harmonic coordinates for character articulation}.
\newblock \bibinfo{journal}{\emph{ACM Trans. Graph.}} \bibinfo{volume}{26}, \bibinfo{number}{3} (\bibinfo{date}{jul} \bibinfo{year}{2007}), \bibinfo{pages}{71–es}.
\newblock
\showISSN{0730-0301}
\urldef\tempurl%
\url{https://doi.org/10.1145/1276377.1276466}
\showDOI{\tempurl}


\bibitem[Kim et~al\mbox{.}(2022)]%
        {Kim_2022_CVPR}
\bibfield{author}{\bibinfo{person}{Gwanghyun Kim}, \bibinfo{person}{Taesung Kwon}, {and} \bibinfo{person}{Jong~Chul Ye}.} \bibinfo{year}{2022}\natexlab{}.
\newblock \showarticletitle{DiffusionCLIP: Text-Guided Diffusion Models for Robust Image Manipulation}. In \bibinfo{booktitle}{\emph{Proceedings of the IEEE/CVF Conference on Computer Vision and Pattern Recognition (CVPR)}}. \bibinfo{pages}{2426--2435}.
\newblock


\bibitem[Kodnongbua et~al\mbox{.}(2023)]%
        {reparam_cad}
\bibfield{author}{\bibinfo{person}{Milin Kodnongbua}, \bibinfo{person}{Benjamin Jones}, \bibinfo{person}{Maaz Bin~Safeer Ahmad}, \bibinfo{person}{Vladimir Kim}, {and} \bibinfo{person}{Adriana Schulz}.} \bibinfo{year}{2023}\natexlab{}.
\newblock \showarticletitle{ReparamCAD: Zero-shot CAD Re-Parameterization for Interactive Manipulation}. In \bibinfo{booktitle}{\emph{SIGGRAPH Asia 2023 Conference Papers}} (<conf-loc>, <city>Sydney</city>, <state>NSW</state>, <country>Australia</country>, </conf-loc>) \emph{(\bibinfo{series}{SA '23})}. \bibinfo{publisher}{Association for Computing Machinery}, \bibinfo{address}{New York, NY, USA}, Article \bibinfo{articleno}{69}, \bibinfo{numpages}{12}~pages.
\newblock
\showISBNx{9798400703157}
\urldef\tempurl%
\url{https://doi.org/10.1145/3610548.3618219}
\showDOI{\tempurl}


\bibitem[Kraevoy et~al\mbox{.}(2008)]%
        {resize_old}
\bibfield{author}{\bibinfo{person}{Vladislav Kraevoy}, \bibinfo{person}{Alla Sheffer}, \bibinfo{person}{Ariel Shamir}, {and} \bibinfo{person}{Daniel Cohen-Or}.} \bibinfo{year}{2008}\natexlab{}.
\newblock \showarticletitle{Non-homogeneous resizing of complex models}.
\newblock \bibinfo{journal}{\emph{ACM Trans. Graph.}} \bibinfo{volume}{27}, \bibinfo{number}{5}, Article \bibinfo{articleno}{111} (\bibinfo{date}{dec} \bibinfo{year}{2008}), \bibinfo{numpages}{9}~pages.
\newblock
\showISSN{0730-0301}
\urldef\tempurl%
\url{https://doi.org/10.1145/1409060.1409064}
\showDOI{\tempurl}


\bibitem[Kulits et~al\mbox{.}(2024)]%
        {kulits2024igllm}
\bibfield{author}{\bibinfo{person}{Peter Kulits}, \bibinfo{person}{Haiwen Feng}, \bibinfo{person}{Weiyang Liu}, \bibinfo{person}{Victoria Abrevaya}, {and} \bibinfo{person}{Michael~J. Black}.} \bibinfo{year}{2024}\natexlab{}.
\newblock \bibinfo{title}{Re-Thinking Inverse Graphics With Large Language Models}.
\newblock
\newblock
\showeprint[arxiv]{2404.xxxxx}~[cs.CL]


\bibitem[Lin et~al\mbox{.}(2011)]%
        {architecture_2}
\bibfield{author}{\bibinfo{person}{Jinjie Lin}, \bibinfo{person}{Daniel Cohen-Or}, \bibinfo{person}{Hao Zhang}, \bibinfo{person}{Cheng Liang}, \bibinfo{person}{Andrei Sharf}, \bibinfo{person}{Oliver Deussen}, {and} \bibinfo{person}{Baoquan Chen}.} \bibinfo{year}{2011}\natexlab{}.
\newblock \showarticletitle{Structure-preserving retargeting of irregular 3D architecture}.
\newblock \bibinfo{journal}{\emph{ACM Trans. Graph.}} \bibinfo{volume}{30}, \bibinfo{number}{6} (\bibinfo{date}{dec} \bibinfo{year}{2011}), \bibinfo{pages}{1–10}.
\newblock
\showISSN{0730-0301}
\urldef\tempurl%
\url{https://doi.org/10.1145/2070781.2024217}
\showDOI{\tempurl}


\bibitem[Lin et~al\mbox{.}(2023)]%
        {lin2023text}
\bibfield{author}{\bibinfo{person}{Yuanze Lin}, \bibinfo{person}{Yi-Wen Chen}, \bibinfo{person}{Yi-Hsuan Tsai}, \bibinfo{person}{Lu Jiang}, {and} \bibinfo{person}{Ming-Hsuan Yang}.} \bibinfo{year}{2023}\natexlab{}.
\newblock \showarticletitle{Text-Driven Image Editing via Learnable Regions}.
\newblock \bibinfo{journal}{\emph{arXiv preprint arXiv:2311.16432}} (\bibinfo{year}{2023}).
\newblock


\bibitem[Makatura et~al\mbox{.}(2023)]%
        {makatura2023large}
\bibfield{author}{\bibinfo{person}{Liane Makatura}, \bibinfo{person}{Michael Foshey}, \bibinfo{person}{Bohan Wang}, \bibinfo{person}{Felix HähnLein}, \bibinfo{person}{Pingchuan Ma}, \bibinfo{person}{Bolei Deng}, \bibinfo{person}{Megan Tjandrasuwita}, \bibinfo{person}{Andrew Spielberg}, \bibinfo{person}{Crystal~Elaine Owens}, \bibinfo{person}{Peter~Yichen Chen}, \bibinfo{person}{Allan Zhao}, \bibinfo{person}{Amy Zhu}, \bibinfo{person}{Wil~J Norton}, \bibinfo{person}{Edward Gu}, \bibinfo{person}{Joshua Jacob}, \bibinfo{person}{Yifei Li}, \bibinfo{person}{Adriana Schulz}, {and} \bibinfo{person}{Wojciech Matusik}.} \bibinfo{year}{2023}\natexlab{}.
\newblock \bibinfo{title}{How Can Large Language Models Help Humans in Design and Manufacturing?}
\newblock
\newblock
\showeprint[arxiv]{2307.14377}~[cs.CL]


\bibitem[Meurer et~al\mbox{.}(2017)]%
        {sympy}
\bibfield{author}{\bibinfo{person}{Aaron Meurer}, \bibinfo{person}{Christopher~P. Smith}, \bibinfo{person}{Mateusz Paprocki}, \bibinfo{person}{Ond\v{r}ej \v{C}ert\'{i}k}, \bibinfo{person}{Sergey~B. Kirpichev}, \bibinfo{person}{Matthew Rocklin}, \bibinfo{person}{AMiT Kumar}, \bibinfo{person}{Sergiu Ivanov}, \bibinfo{person}{Jason~K. Moore}, \bibinfo{person}{Sartaj Singh}, \bibinfo{person}{Thilina Rathnayake}, \bibinfo{person}{Sean Vig}, \bibinfo{person}{Brian~E. Granger}, \bibinfo{person}{Richard~P. Muller}, \bibinfo{person}{Francesco Bonazzi}, \bibinfo{person}{Harsh Gupta}, \bibinfo{person}{Shivam Vats}, \bibinfo{person}{Fredrik Johansson}, \bibinfo{person}{Fabian Pedregosa}, \bibinfo{person}{Matthew~J. Curry}, \bibinfo{person}{Andy~R. Terrel}, \bibinfo{person}{\v{S}t\v{e}p\'{a}n Rou\v{c}ka}, \bibinfo{person}{Ashutosh Saboo}, \bibinfo{person}{Isuru Fernando}, \bibinfo{person}{Sumith Kulal}, \bibinfo{person}{Robert Cimrman}, {and} \bibinfo{person}{Anthony Scopatz}.} \bibinfo{year}{2017}\natexlab{}.
\newblock \showarticletitle{{SymPy}: symbolic computing in Python}.
\newblock \bibinfo{journal}{\emph{PeerJ Computer Science}}  \bibinfo{volume}{3} (\bibinfo{date}{Jan.} \bibinfo{year}{2017}), \bibinfo{pages}{e103}.
\newblock
\showISSN{2376-5992}
\urldef\tempurl%
\url{https://doi.org/10.7717/peerj-cs.103}
\showDOI{\tempurl}


\bibitem[Michel and Boubekeur(2021)]%
        {dag_amendments}
\bibfield{author}{\bibinfo{person}{\'{E}lie Michel} {and} \bibinfo{person}{Tamy Boubekeur}.} \bibinfo{year}{2021}\natexlab{}.
\newblock \showarticletitle{DAG amendment for inverse control of parametric shapes}.
\newblock \bibinfo{journal}{\emph{ACM Trans. Graph.}} \bibinfo{volume}{40}, \bibinfo{number}{4}, Article \bibinfo{articleno}{173} (\bibinfo{date}{jul} \bibinfo{year}{2021}), \bibinfo{numpages}{14}~pages.
\newblock
\showISSN{0730-0301}
\urldef\tempurl%
\url{https://doi.org/10.1145/3450626.3459823}
\showDOI{\tempurl}


\bibitem[Michel et~al\mbox{.}(2021)]%
        {text2mesh}
\bibfield{author}{\bibinfo{person}{Oscar Michel}, \bibinfo{person}{Roi Bar-On}, \bibinfo{person}{Richard Liu}, \bibinfo{person}{Sagie Benaim}, {and} \bibinfo{person}{Rana Hanocka}.} \bibinfo{year}{2021}\natexlab{}.
\newblock \showarticletitle{Text2Mesh: Text-Driven Neural Stylization for Meshes}.
\newblock \bibinfo{journal}{\emph{arXiv preprint arXiv:2112.03221}} (\bibinfo{year}{2021}).
\newblock


\bibitem[Mitra et~al\mbox{.}(2013)]%
        {niloy_survey}
\bibfield{author}{\bibinfo{person}{Niloy Mitra}, \bibinfo{person}{Michael Wand}, \bibinfo{person}{Hao~(Richard) Zhang}, \bibinfo{person}{Daniel Cohen-Or}, \bibinfo{person}{Vladimir Kim}, {and} \bibinfo{person}{Qi-Xing Huang}.} \bibinfo{year}{2013}\natexlab{}.
\newblock \showarticletitle{Structure-aware shape processing}. In \bibinfo{booktitle}{\emph{SIGGRAPH Asia 2013 Courses}} (Hong Kong, Hong Kong) \emph{(\bibinfo{series}{SA '13})}. \bibinfo{publisher}{Association for Computing Machinery}, \bibinfo{address}{New York, NY, USA}, Article \bibinfo{articleno}{1}, \bibinfo{numpages}{20}~pages.
\newblock
\showISBNx{9781450326315}
\urldef\tempurl%
\url{https://doi.org/10.1145/2542266.2542267}
\showDOI{\tempurl}


\bibitem[Mo et~al\mbox{.}(2019a)]%
        {mo2019structedit}
\bibfield{author}{\bibinfo{person}{Kaichun Mo}, \bibinfo{person}{Paul Guerrero}, \bibinfo{person}{Li Yi}, \bibinfo{person}{Hao Su}, \bibinfo{person}{Peter Wonka}, \bibinfo{person}{Niloy Mitra}, {and} \bibinfo{person}{Leonidas~J Guibas}.} \bibinfo{year}{2019}\natexlab{a}.
\newblock \showarticletitle{{StructEdit}: Learning Structural Shape Variations}.
\newblock \bibinfo{journal}{\emph{Arxiv1911.11098}} (\bibinfo{year}{2019}).
\newblock


\bibitem[Mo et~al\mbox{.}(2019b)]%
        {partnet}
\bibfield{author}{\bibinfo{person}{Kaichun Mo}, \bibinfo{person}{Shilin Zhu}, \bibinfo{person}{Angel~X. Chang}, \bibinfo{person}{Li Yi}, \bibinfo{person}{Subarna Tripathi}, \bibinfo{person}{Leonidas~J. Guibas}, {and} \bibinfo{person}{Hao Su}.} \bibinfo{year}{2019}\natexlab{b}.
\newblock \showarticletitle{{PartNet}: A Large-Scale Benchmark for Fine-Grained and Hierarchical Part-Level {3D} Object Understanding}. In \bibinfo{booktitle}{\emph{CVPR}}.
\newblock


\bibitem[OpenAI(2024)]%
        {openai2024gpt4}
\bibfield{author}{\bibinfo{person}{OpenAI}.} \bibinfo{year}{2024}\natexlab{}.
\newblock \bibinfo{title}{GPT-4 Technical Report}.
\newblock
\newblock
\showeprint[arxiv]{2303.08774}~[cs.CL]


\bibitem[Sederberg and Parry(1986)]%
        {Sederberg}
\bibfield{author}{\bibinfo{person}{Thomas~W. Sederberg} {and} \bibinfo{person}{Scott~R. Parry}.} \bibinfo{year}{1986}\natexlab{}.
\newblock \showarticletitle{Free-form deformation of solid geometric models}. In \bibinfo{booktitle}{\emph{Proceedings of the 13th Annual Conference on Computer Graphics and Interactive Techniques}} \emph{(\bibinfo{series}{SIGGRAPH '86})}. \bibinfo{publisher}{Association for Computing Machinery}, \bibinfo{address}{New York, NY, USA}, \bibinfo{pages}{151–160}.
\newblock
\showISBNx{0897911962}
\urldef\tempurl%
\url{https://doi.org/10.1145/15922.15903}
\showDOI{\tempurl}


\bibitem[Shtof et~al\mbox{.}(2013)]%
        {geo_snap}
\bibfield{author}{\bibinfo{person}{Alex Shtof}, \bibinfo{person}{Alexander Agathos}, \bibinfo{person}{Yotam Gingold}, \bibinfo{person}{Ariel Shamir}, {and} \bibinfo{person}{Daniel Cohen-Or}.} \bibinfo{year}{2013}\natexlab{}.
\newblock \showarticletitle{Geosemantic Snapping for Sketch-Based Modeling}.
\newblock \bibinfo{journal}{\emph{Computer Graphics Forum}} \bibinfo{volume}{32}, \bibinfo{number}{2} (\bibinfo{year}{2013}), \bibinfo{pages}{245--253}.
\newblock
\showISSN{1467-8659}
\urldef\tempurl%
\url{https://doi.org/10.1111/cgf.12044}
\showDOI{\tempurl}
\newblock
\shownote{Proceedings of Eurographics 2013}.


\bibitem[Singh and Fiume(1998)]%
        {orig_wires}
\bibfield{author}{\bibinfo{person}{Karan Singh} {and} \bibinfo{person}{Eugene Fiume}.} \bibinfo{year}{1998}\natexlab{}.
\newblock \showarticletitle{Wires: a geometric deformation technique}. In \bibinfo{booktitle}{\emph{Proceedings of the 25th Annual Conference on Computer Graphics and Interactive Techniques}} \emph{(\bibinfo{series}{SIGGRAPH '98})}. \bibinfo{publisher}{Association for Computing Machinery}, \bibinfo{address}{New York, NY, USA}, \bibinfo{pages}{405–414}.
\newblock
\showISBNx{0897919998}
\urldef\tempurl%
\url{https://doi.org/10.1145/280814.280946}
\showDOI{\tempurl}


\bibitem[Slim and Elhoseiny(2024)]%
        {shapewalk}
\bibfield{author}{\bibinfo{person}{Habib Slim} {and} \bibinfo{person}{Mohamed Elhoseiny}.} \bibinfo{year}{2024}\natexlab{}.
\newblock \showarticletitle{{ShapeWalk}: Compositional Shape Editing through Language-Guided Chains}. In \bibinfo{booktitle}{\emph{Conference on Computer Vision and Pattern Recognition (CVPR)}}.
\newblock


\bibitem[Slim et~al\mbox{.}(2023)]%
        {slim2023_3dcompatplus}
\bibfield{author}{\bibinfo{person}{Habib Slim}, \bibinfo{person}{Xiang Li}, \bibinfo{person}{Yuchen Li}, \bibinfo{person}{Mahmoud Ahmed}, \bibinfo{person}{Mohamed Ayman}, \bibinfo{person}{Ujjwal Upadhyay}, \bibinfo{person}{Ahmed Abdelreheem}, \bibinfo{person}{Arpit Prajapati}, \bibinfo{person}{Suhail Pothigara}, \bibinfo{person}{Peter Wonka}, {and} \bibinfo{person}{Mohamed Elhoseiny}.} \bibinfo{year}{2023}\natexlab{}.
\newblock \showarticletitle{3DCoMPaT++: An improved Large-scale 3D Vision Dataset for Compositional Recognition}.
\newblock  (\bibinfo{year}{2023}).
\newblock


\bibitem[Sorkine and Alexa(2007)]%
        {ARAP_modeling}
\bibfield{author}{\bibinfo{person}{Olga Sorkine} {and} \bibinfo{person}{Marc Alexa}.} \bibinfo{year}{2007}\natexlab{}.
\newblock \showarticletitle{As-Rigid-As-Possible Surface Modeling}. In \bibinfo{booktitle}{\emph{Proceedings of EUROGRAPHICS/ACM SIGGRAPH Symposium on Geometry Processing}}. \bibinfo{pages}{109--116}.
\newblock


\bibitem[Sumner et~al\mbox{.}(2007)]%
        {Sumner}
\bibfield{author}{\bibinfo{person}{Robert~W. Sumner}, \bibinfo{person}{Johannes Schmid}, {and} \bibinfo{person}{Mark Pauly}.} \bibinfo{year}{2007}\natexlab{}.
\newblock \showarticletitle{Embedded deformation for shape manipulation}.
\newblock \bibinfo{journal}{\emph{ACM Trans. Graph.}} \bibinfo{volume}{26}, \bibinfo{number}{3} (\bibinfo{date}{jul} \bibinfo{year}{2007}), \bibinfo{pages}{80–es}.
\newblock
\showISSN{0730-0301}
\urldef\tempurl%
\url{https://doi.org/10.1145/1276377.1276478}
\showDOI{\tempurl}


\bibitem[Sur\'is et~al\mbox{.}(2023)]%
        {surismenon2023vipergpt}
\bibfield{author}{\bibinfo{person}{D\'idac Sur\'is}, \bibinfo{person}{Sachit Menon}, {and} \bibinfo{person}{Carl Vondrick}.} \bibinfo{year}{2023}\natexlab{}.
\newblock \showarticletitle{ViperGPT: Visual Inference via Python Execution for Reasoning}.
\newblock \bibinfo{journal}{\emph{Proceedings of IEEE International Conference on Computer Vision (ICCV)}} (\bibinfo{year}{2023}).
\newblock


\bibitem[Wang et~al\mbox{.}(2023b)]%
        {wang2023mdp}
\bibfield{author}{\bibinfo{person}{Qian Wang}, \bibinfo{person}{Biao Zhang}, \bibinfo{person}{Michael Birsak}, {and} \bibinfo{person}{Peter Wonka}.} \bibinfo{year}{2023}\natexlab{b}.
\newblock \bibinfo{title}{MDP: A Generalized Framework for Text-Guided Image Editing by Manipulating the Diffusion Path}.
\newblock
\newblock
\showeprint[arxiv]{2303.16765}~[cs.CV]


\bibitem[Wang et~al\mbox{.}(2023a)]%
        {voting}
\bibfield{author}{\bibinfo{person}{Xuezhi Wang}, \bibinfo{person}{Jason Wei}, \bibinfo{person}{Dale Schuurmans}, \bibinfo{person}{Quoc~V Le}, \bibinfo{person}{Ed~H. Chi}, \bibinfo{person}{Sharan Narang}, \bibinfo{person}{Aakanksha Chowdhery}, {and} \bibinfo{person}{Denny Zhou}.} \bibinfo{year}{2023}\natexlab{a}.
\newblock \showarticletitle{Self-Consistency Improves Chain of Thought Reasoning in Language Models}. In \bibinfo{booktitle}{\emph{The Eleventh International Conference on Learning Representations}}.
\newblock
\urldef\tempurl%
\url{https://openreview.net/forum?id=1PL1NIMMrw}
\showURL{%
\tempurl}


\bibitem[Wang et~al\mbox{.}(2011)]%
        {wang2011}
\bibfield{author}{\bibinfo{person}{Yanzhen Wang}, \bibinfo{person}{Kai Xu}, \bibinfo{person}{Jun Li}, \bibinfo{person}{Hao Zhang}, \bibinfo{person}{Ariel Shamir}, \bibinfo{person}{Ligang Liu}, \bibinfo{person}{Zhiquan Cheng}, {and} \bibinfo{person}{Yueshan Xiong}.} \bibinfo{year}{2011}\natexlab{}.
\newblock \showarticletitle{Symmetry Hierarchy of Man-Made Objects}.
\newblock \bibinfo{journal}{\emph{Comput. Graph. Forum}} (\bibinfo{year}{2011}).
\newblock


\bibitem[Wei et~al\mbox{.}(2023)]%
        {wei2023chainofthought}
\bibfield{author}{\bibinfo{person}{Jason Wei}, \bibinfo{person}{Xuezhi Wang}, \bibinfo{person}{Dale Schuurmans}, \bibinfo{person}{Maarten Bosma}, \bibinfo{person}{Brian Ichter}, \bibinfo{person}{Fei Xia}, \bibinfo{person}{Ed Chi}, \bibinfo{person}{Quoc Le}, {and} \bibinfo{person}{Denny Zhou}.} \bibinfo{year}{2023}\natexlab{}.
\newblock \showarticletitle{Chain-of-Thought Prompting Elicits Reasoning in Large Language Models}. In \bibinfo{booktitle}{\emph{NeurIPS}}.
\newblock


\bibitem[Xu et~al\mbox{.}(2009)]%
        {joint_deform}
\bibfield{author}{\bibinfo{person}{Weiwei Xu}, \bibinfo{person}{Jun Wang}, \bibinfo{person}{KangKang Yin}, \bibinfo{person}{Kun Zhou}, \bibinfo{person}{Michiel van~de Panne}, \bibinfo{person}{Falai Chen}, {and} \bibinfo{person}{Baining Guo}.} \bibinfo{year}{2009}\natexlab{}.
\newblock \showarticletitle{Joint-aware manipulation of deformable models}.
\newblock \bibinfo{journal}{\emph{ACM Trans. Graph.}} \bibinfo{volume}{28}, \bibinfo{number}{3}, Article \bibinfo{articleno}{35} (\bibinfo{date}{jul} \bibinfo{year}{2009}), \bibinfo{numpages}{9}~pages.
\newblock
\showISSN{0730-0301}
\urldef\tempurl%
\url{https://doi.org/10.1145/1531326.1531341}
\showDOI{\tempurl}


\bibitem[Yamada et~al\mbox{.}(2024)]%
        {yamada2024l3go}
\bibfield{author}{\bibinfo{person}{Yutaro Yamada}, \bibinfo{person}{Khyathi Chandu}, \bibinfo{person}{Yuchen Lin}, \bibinfo{person}{Jack Hessel}, \bibinfo{person}{Ilker Yildirim}, {and} \bibinfo{person}{Yejin Choi}.} \bibinfo{year}{2024}\natexlab{}.
\newblock \bibinfo{title}{L3GO: Language Agents with Chain-of-3D-Thoughts for Generating Unconventional Objects}.
\newblock
\newblock
\showeprint[arxiv]{2402.09052}~[cs.AI]


\bibitem[Yang et~al\mbox{.}(2023)]%
        {yang2023holodeck}
\bibfield{author}{\bibinfo{person}{Yue Yang}, \bibinfo{person}{Fan-Yun Sun}, \bibinfo{person}{Luca Weihs}, \bibinfo{person}{Eli VanderBilt}, \bibinfo{person}{Alvaro Herrasti}, \bibinfo{person}{Winson Han}, \bibinfo{person}{Jiajun Wu}, \bibinfo{person}{Nick Haber}, \bibinfo{person}{Ranjay Krishna}, \bibinfo{person}{Lingjie Liu}, \bibinfo{person}{Chris Callison-Burch}, \bibinfo{person}{Mark Yatskar}, \bibinfo{person}{Aniruddha Kembhavi}, {and} \bibinfo{person}{Christopher Clark}.} \bibinfo{year}{2023}\natexlab{}.
\newblock \showarticletitle{Holodeck: Language Guided Generation of 3D Embodied AI Environments}.
\newblock \bibinfo{journal}{\emph{arXiv preprint arXiv:2312.09067}} (\bibinfo{year}{2023}).
\newblock


\bibitem[Yumer et~al\mbox{.}(2015)]%
        {yumer_semantic}
\bibfield{author}{\bibinfo{person}{Mehmet~Ersin Yumer}, \bibinfo{person}{Siddhartha Chaudhuri}, \bibinfo{person}{Jessica~K. Hodgins}, {and} \bibinfo{person}{Levent~Burak Kara}.} \bibinfo{year}{2015}\natexlab{}.
\newblock \showarticletitle{Semantic shape editing using deformation handles}.
\newblock \bibinfo{journal}{\emph{ACM Trans. Graph.}} \bibinfo{volume}{34}, \bibinfo{number}{4}, Article \bibinfo{articleno}{86} (\bibinfo{date}{jul} \bibinfo{year}{2015}), \bibinfo{numpages}{12}~pages.
\newblock
\showISSN{0730-0301}
\urldef\tempurl%
\url{https://doi.org/10.1145/2766908}
\showDOI{\tempurl}


\bibitem[Zheng et~al\mbox{.}(2011)]%
        {comp_prop}
\bibfield{author}{\bibinfo{person}{Youyi Zheng}, \bibinfo{person}{Hongbo Fu}, \bibinfo{person}{Daniel Cohen-Or}, \bibinfo{person}{Oscar Kin-Chung Au}, {and} \bibinfo{person}{Chiew-Lan Tai}.} \bibinfo{year}{2011}\natexlab{}.
\newblock \showarticletitle{Component-wise Controllers for Structure-Preserving Shape Manipulation}.
\newblock \bibinfo{journal}{\emph{Computer Graphics Forum}} \bibinfo{volume}{30}, \bibinfo{number}{2} (\bibinfo{year}{2011}), \bibinfo{pages}{563--572}.
\newblock
\urldef\tempurl%
\url{https://doi.org/10.1111/j.1467-8659.2011.01880.x}
\showDOI{\tempurl}
\showeprint{https://onlinelibrary.wiley.com/doi/pdf/10.1111/j.1467-8659.2011.01880.x}


\bibitem[Zhou et~al\mbox{.}(2023)]%
        {zhou2023partslip}
\bibfield{author}{\bibinfo{person}{Yuchen Zhou}, \bibinfo{person}{Jiayuan Gu}, \bibinfo{person}{Xuanlin Li}, \bibinfo{person}{Minghua Liu}, \bibinfo{person}{Yunhao Fang}, {and} \bibinfo{person}{Hao Su}.} \bibinfo{year}{2023}\natexlab{}.
\newblock \bibinfo{title}{PartSLIP++: Enhancing Low-Shot 3D Part Segmentation via Multi-View Instance Segmentation and Maximum Likelihood Estimation}.
\newblock
\newblock
\showeprint[arxiv]{2312.03015}~[cs.CV]


\end{thebibliography}

\end{document}